\documentclass[runningheads]{llncs}

\usepackage{eccv}

\usepackage{eccvabbrv}

\usepackage{graphicx}
\usepackage{booktabs}
\usepackage{multirow}
\usepackage{graphicx}
\usepackage[accsupp]{axessibility}  
\usepackage{overpic}
\usepackage{bbding,pifont}
\usepackage{cite}

\usepackage[pagebackref,breaklinks,colorlinks,citecolor=eccvblue]{hyperref}

\usepackage{orcidlink}
\usepackage{amsmath}
\usepackage{bm}
\usepackage{appendix}

\newcommand{\mat}[1]{\bm{\mathit{#1}}}
\newcommand{\ii}[1]{\mathit{#1}}
\newcommand{\bb}[1]{\mathbf{#1}}

\begin{document}
	
	\title{GLAD: Towards Better Reconstruction with Global and Local Adaptive Diffusion Models for Unsupervised Anomaly Detection}
	
	\titlerunning{GLAD}
	
	\author{Hang Yao\inst{1} \and
		Ming Liu\inst{1,2}$^{(}$\Envelope$^)$ \and
		Haolin Wang\inst{1} \and
		Zhicun Yin\inst{1} \and \\
		Zifei Yan\inst{1,2} \and
		Xiaopeng Hong\inst{1} \and
		Wangmeng Zuo\inst{1,2}\\
		\email{\href{mailto:yaohang_1@outlook.com}{yaohang\_1@outlook.com}, \href{mailto:csmliu@outlook.com}{csmliu@outlook.com}, \href{mailto:cswmzuo@gmail.com}{cswmzuo@gmail.com}}}
	
	\authorrunning{H.~Yao~\etal}
	
	\institute{$^1$Harbin Institute of Technology, Harbin, China\\$^2$Pazhou Lab Huangpu, Guangzhou, China}
	
	\maketitle

	\begin{abstract}
		
		Diffusion models have shown superior performance on unsupervised anomaly detection tasks. Since trained with normal data only, diffusion models tend to reconstruct normal counterparts of test images with certain noises added.
		However, these methods treat all potential anomalies equally, which may cause two main problems.
		From the global perspective, the difficulty of reconstructing images with different anomalies is uneven.
		For example, adding back a missing element is harder than dealing with a scratch, thus requiring a larger number of denoising steps.
		Therefore, instead of utilizing the same setting for all samples, we propose to predict a particular denoising step for each sample by evaluating the difference between image contents and the priors extracted from diffusion models.
		From the local perspective, reconstructing abnormal regions differs from normal areas even in the same image.
		Theoretically, the diffusion model predicts a noise for each step, typically following a standard Gaussian distribution.
		However, due to the difference between the anomaly and its potential normal counterpart, the predicted noise in abnormal regions will inevitably deviate from the standard Gaussian distribution.
		To this end, we propose introducing synthetic abnormal samples in training to encourage the diffusion models to break through the limitation of standard Gaussian distribution, and a spatial-adaptive feature fusion scheme is utilized during inference.
		With the above modifications, we propose a global and local adaptive diffusion model (abbreviated to GLAD) for unsupervised anomaly detection, which introduces appealing flexibility and achieves anomaly-free reconstruction while retaining as much normal information as possible.
		Extensive experiments are conducted on three commonly used anomaly detection datasets (MVTec-AD, MPDD, and VisA) and a printed circuit board dataset (PCB-Bank) we integrated, showing the effectiveness of the proposed method.
		The source code and pre-trained models are publicly available at \url{https://github.com/hyao1/GLAD}.
	\end{abstract}
	
	\section{Introduction}
	\label{sec:intro}
	
	Anomaly detection (AD) aims to detect and locate abnormal patterns that influence the appearance and function of objects, which is vital for the quality of products and has been widely used in industries~\cite{SPD20,Simplenet23,APRILGAN18}.
	In practice, the prevalence of different anomaly types varies, making it challenging to collect enough abnormal samples for all anomaly types in situations with high yield rates.
	Furthermore, due to the ever-changing product design and production processes, it is impossible to collect all anomalies in advance.
	Therefore, unsupervised anomaly detection (UAD) has drawn much attention with only normal samples required.
	To achieve unsupervised anomaly detection, reconstruction-based methods generate a potential normal sample corresponding to the given one, and the anomalies can be detected and located via the comparison between the given sample and its normal counterpart.
	Due to the prominent modeling ability, diffusion models are introduced for sample reconstruction and have shown superior performance.
	
	
	Existing diffusion model-based UAD methods~\cite{DiAD37,26,Lafite38,DDAD8} typically follow a common process to reconstruct the test samples.
	To begin with, a diffusion model is trained with normal samples of certain objects or products (\eg, bottles, hazelnuts, \etc).
	Then, it can be deployed to reconstruct a sample with random noise added.
	Note that during the training process, the diffusion model captures the distribution of normal samples only, which implies that it will generate a normal sample from any noise-contaminated inputs as long as the randomness is strong enough\footnote{In the setting of diffusion models, the randomness is equivalent to the weight of the random noise, which is determined by the denoising step. In other words, a larger denoising step means higher noise weight and stronger randomness.}.
	Therefore, existing methods choose to set a sufficiently large denoising step to guarantee the reconstruction ability.
	
	\begin{figure}[t]
		\centering
		\includegraphics[width=.99\linewidth]{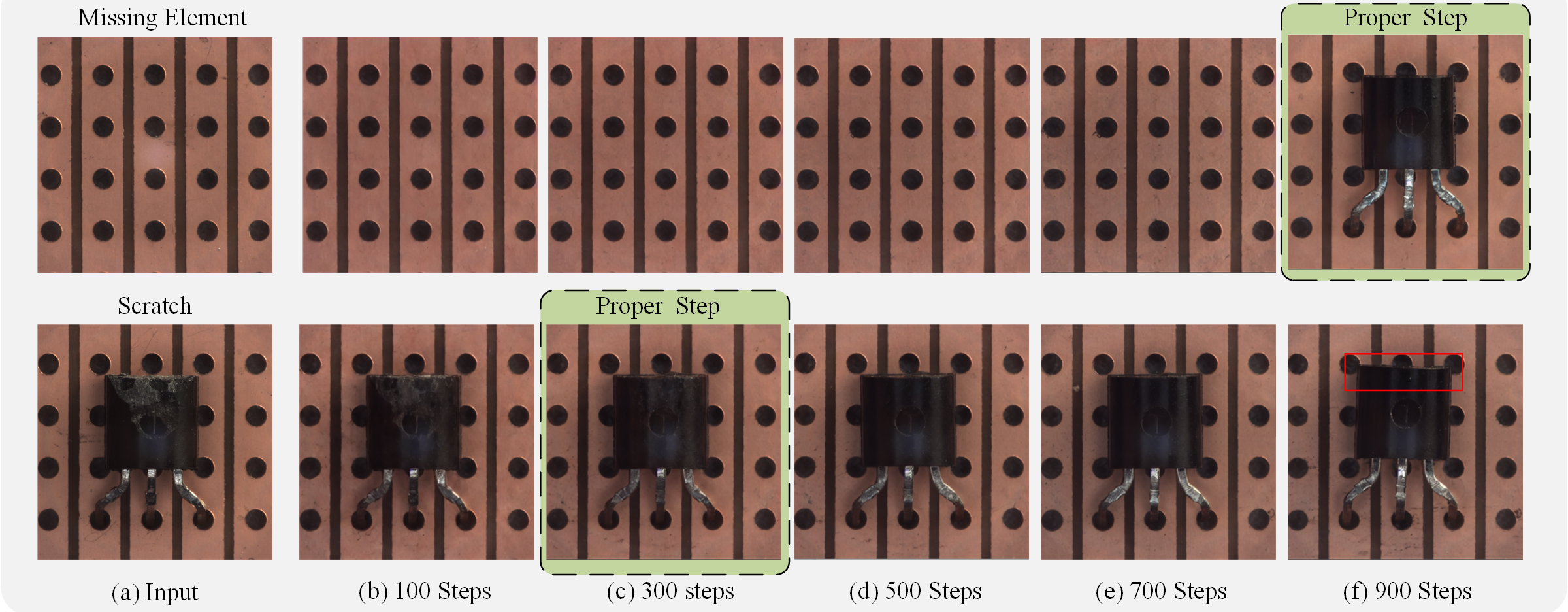}
		\caption{Illustration of adaptive denoising process. For severe anomalies like missing elements, it requires a large number of denoising steps (900) to add the element back, while for small anomalies like scratch, 300 steps are already enough. Besides, setting a large enough denoising step (\eg, 900) for all samples will affect the detail preservation. For example, in the area bounded by red lines, the position of the element is changed, which will be marked as anomalies during the comparison process.}\label{fig:denoising_steps}
	\end{figure}
	
	However, setting the same denoising step for all samples is a sub-optimal solution.
	As shown in \cref{fig:denoising_steps}, the difficulty of reconstructing images with different anomalies is uneven. For example, 900 steps are required to add a missing element back, while 300 steps are already enough to deal with a scratch. Besides, apart from better reconstruction ability, a larger denoising step also means higher randomness and uncertainty, leading to less preserved details of the original test samples (see the areas bounded by the red lines in \cref{fig:denoising_steps}.) To this end, we propose to set an Adaptive Denoising Step (ADP) for each sample, which achieves a better trade-off between reconstruction quality and detail preservation ability. In order to implement such an adaptive denoising step method, we take advantage of the prior in the diffusion models. Specifically, we first add noise to the test sample with a large enough noise weight, and perform the denoising steps to gradually remove the noises and reconstruct a normal sample. During the reconstruction procedure, we can compare the reconstructed sample with the noise-contaminated input, where the difference reflects the existence of anomalies and can help adaptively determine the denoising steps. Since this proper steps is used to add noise to the whole image, so we regard it as a global adaptive setting. 
	
	Apart from the global denoising step, from a local perspective, we also find that even in the same image, reconstructing the abnormal regions is different from the normal areas.
	For normal areas, the diffusion model only needs to remove the added noise. It means that the noise to be predicted is exactly the one added, both following the standard Gaussian distribution.
	However, for abnormal regions, in order to reconstruct their normal counterparts, the noise to be predicted by the diffusion model inevitably deviates from the standard Gaussian distribution, making the prediction more difficult than in the normal areas.
	As a remedy, we propose Anomaly-oriented Training Paradigm (ATP), which introduces synthetic abnormal samples when training the diffusion models and generalizes the loss function of the diffusion model to a more general form. The proposed Anomaly-Oriented Training Paradigm encourages the diffusion models to break through the limitation of standard Gaussian distribution and promotes the ability to generate normal samples in the abnormal regions.
	Besides, for better detail preservation, we introduce an Spatial-Adaptive Feature Fusion (SAFF) scheme during inference by fusing sample features of the normal regions and generated features in abnormal regions, which better preserves the details in potential normal regions and alleviates the difficulty of the subsequent comparison procedure. 
	In this way, the diffusion model becomes flexible enough to achieve a local adaptive inference for different regions (normal and abnormal regions), with both reconstruction and detail preservation abilities equipped.
	
	With the above two modifications, a global and local adaptive diffusion model (abbreviated to GLAD) is presented.
	Extensive experiments are conducted on four anomaly detection datasets (\ie, MVTec-AD~\cite{MVTec-AD28}, MPDD~\cite{MPDD29}, VisA~\cite{visa40}, and the PCB-Bank dataset\footnote{PCB-Bank is a printed circuit board dataset we integrated from existing datasets, please refer to \url{https://github.com/SSRheart/industrial-anomaly-detection-dataset} for more details.}) to verify the effectiveness of the proposed method.
	The experimental results show that our method achieves superior performance on unsupervised anomaly detection tasks.
	The contributions of this paper are as follows.
	
	\begin{itemize}
		\item For a better trade-off between reconstruction quality and detail preservation ability, unlike existing diffusion model-based methods utilizing the same setting for all samples, we propose to predict an Adaptive Denoising Step (ADP) as a global adaptive setting for each sample to retain more normal information.
		\item Considering the difference between abnormal regions and normal areas, we introduce Anomaly-oriented Training Paradigm (ATP) during training to allow diffusion model to predict non-Gaussian noise at abnormal regions, and propose a Spatial-Adaptive Feature Fusion (SAFF) scheme during inference to avoid reconstruction of abnormal regions.
		\item A printed circuit board dataset (PCB-Bank) is integrated for a comprehensive evaluation on PCB products. The experiments on three commonly used datasets and our integrated PCB-Bank show that the proposed global and local adaptive diffusion model (GLAD) improves both reconstruction quality and anomaly detection ability.
	\end{itemize}
	
	\section{Related Work}
	\subsection{Anomaly Detection}
	Mainstream unsupervised anomaly detection methods can be divided into two categories, \ie reconstruction-based methods and embedding-based methods. 
	
	Reconstruction-based methods are based on the hypothesis that models trained on normal samples only can reconstruct normal images well, but can not reconstruct abnormal areas. Anomalies can be detected by comparing the samples before and after reconstruction. Early methods~\cite{2,STAE3,AD-FactorVAE4} attempt to utilize variational auto-encoders~\cite{vae1} to reconstruct samples. OCR-GAN~\cite{OCR-GAN5} decouples images into different frequency components and models the reconstruction process as a combination of parallel omni-frequency image restorations. Some other works explore a broader self-supervised learning paradigm for anomaly reconstruction. UniAD~\cite{UniAD6} uses transformers to reconstruct sample features with masked self-attention and proposes a feature jittering strategy to address the shortcut issue. DRAEM~\cite{DRAEM7} synthesizes pseudo-anomaly images to train a UNet architecture to map abnormal images to normal images. Then a discriminator is also trained with pseudo-anomaly images and reconstructed images to discriminate anomalous areas. Recently, diffusion models~\cite{DDPM16,DDIM17} are proposed and achieve state-of-the-art performance. Lu~\etal~\cite{26} introduces noise to overwhelm the anomalous pixels and obtains pixel-wise precise anomaly scores from the intermediate denoising process. DiffAD~\cite{DiffAD9} diffuses the latent representation of the test image as noisy condition embedding, which contributes to produce high-quality reconstructed images while retaining normal information of test samples. DDAD~\cite{DDAD8} uses score-based function to reintegrate the information of test samples during the denoising process. However, these methods add fixed steps of noise for denoiseing, which is not suitable for various anomaly types.
	
	Embedding-based methods extract feature of images to evaluate abnormal areas. Knowledge distillation-based methods~\cite{10,AST11} first train student network with normal samples. Then, features from the pre-trained teacher network are compared with features from the student network to detect and locate anomalies. Reverse distillation~\cite{RD4AD14} is developed to utilize different architectures of teacher and student to maintain the distinction of anomaly. PaDiM~\cite{Padim13} builds multivariate Gaussian distributions for patch features of normal samples and uses Mahalanobis distance as the anomaly score. PatchCore~\cite{PatchCore24} proposes a memory bank to save features of normal images, which are compared with feature maps of test images to distinguish the difference between normal and abnormal features. 
	
	\subsection{Diffusion Model}
	Inspired by principles of nonequilibrium thermodynamics~\cite{15}, diffusion model (DM)~\cite{DDPM16} is proposed for image generation, which has been utilized in a variety of downstream tasks~\cite{smartcontrol,elite,controlvideo}. Denoising diffusion implicit models (DDIM)~\cite{DDIM17} considers the the reverse process of DM as non-Markovian processes, which speeds up the inference greatly. Latent diffusion model (LDM)~\cite{SD30} conducts training and inference in latent space with pre-trained VAE, further reducing the cost of resources and time. Besides, Text inversion~\cite{TextInversion35} and Dreambooth~\cite{Dreambooth36} learn the appearance of subjects in a given reference set and synthesize novel renditions of them in different contexts. These methods follow the training paradigm for predicting Gaussian noise and start denoising from a Gaussian noise. Thus, these methods can not achieve adaptive denoising to reconstruct a abnormal image into a normal one.
	
	\section{Methodology}
	In this section, we start with the common practice of existing diffusion model-based unsupervised anomaly detection methods.
	By transforming their working processes into formal expressions of formulas, we naturally reveal the existing problems and discover the corresponding solutions, which derive the global and local adaptive diffusion model (\ie, GLAD) in this paper.
	
	\subsection{Preliminary}
	First, we provide preliminary knowledge of diffusion models for later analyses.
	
	\noindent\textbf{Diffusion Process.}
	In the diffusion process, a random noise $\epsilon$ is added to the sample $\mat{x}$, and the result after $\ii{t}$ steps can be represented by,
	\begin{equation}
		\mat{x}_\ii{t} = \sqrt{\bar{\alpha}}_t\mat{x}_0 + \sqrt{1-\bar{\alpha}_\ii{t}}\epsilon,
		\label{eqn:diffusion}
	\end{equation}
	where $\bar{\alpha}_\ii{t}$ is manually defined, which is negatively correlated with $\ii{t}$, and $\mat{x}_0=\mat{x}$.
	
	\noindent\textbf{Intermediate Result Visualization.}
	\cref{eqn:diffusion} can be rewritten to obtain the noise-free version of the intermediate result at the $\ii{t}$-th step, \ie,
	\begin{equation}
		\mat{x}_{\ii{t}\to0}=\tfrac{1}{\sqrt{\bar{\alpha}_\ii{t}}}(\mat{x}_\ii{t}-\sqrt{1-\bar{\alpha}_\ii{t}}\epsilon_\theta(\mat{x}_\ii{t},\ii{t})),
		\label{eqn:step_visualization}
	\end{equation}
	where $\epsilon_\theta$ is from the pre-trained diffusion model for predicting the noise added.
	
	\noindent\textbf{Generation Process.}
	Each step of the generation stage can be formulated by,
	\begin{equation}
		\hat{\mat{x}}_{\ii{t}-1}=
		\sqrt{\bar{\alpha}_{\ii{t}-1}}\hat{\mat{x}}_{\ii{t}\to0}+\sqrt{1-\bar{\alpha}_{\ii{t}-1}}\epsilon_\theta(\hat{\mat{x}}_\ii{t},\ii{t}).
		\label{eqn:generation_step}
	\end{equation}
	Note that the variables predicted by the diffusion model are marked by $\wedge$, for example, $\mat{x}_\ii{t}$ is obtained by adding random noise directly (the diffusion process), while $\hat{\mat{x}}_\ii{t}$ is obtained by denoising from a larger step (the generation process).

	\subsection{Formulaic Analysis on Reconstruction Errors.}
	\label{sec:method_analysis}
	In existing diffusion model-based unsupervised anomaly detection methods, the common way of reconstructing the normal counterpart of a test sample is to add certain noise to the given test sample and then execute the generation process of the diffusion model.
	Denote the test sample with anomalies by $\mat{x}^\ii{a}$, its potential normal counter part by $\mat{x}$, then the process can be described by $\mat{x}^\ii{a}\xrightarrow{\ii{diff}}\mat{x}^\ii{a}_\ii{t}\xrightarrow{\ii{gen}}\hat{\mat{x}}^\ii{a}$, and ideally we should have $\hat{\mat{x}}^\ii{a}\to\mat{x}$.
	Since the anomalies are typically detected and located by comparison between $\hat{\mat{x}}^\ii{a}$ and $\mat{x}^\ii{a}$, we can require that $\|\hat{\mat{x}}^\ii{a}-\mat{x}\|_\infty<\tau$, where $\tau$ is a threshold manually set for distinguishing between normal and abnormal samples.
	
	Since the diffusion model is pre-trained and fixed during the generation process, we analyze in the $\ii{t}$ step.
	Denote the difference between $\mat{x}^\ii{a}$ and $\mat{x}$ by $\mat{n}$, \ie, $\mat{x}^\ii{a}=\mat{x}+\mat{n}$, and according to \cref{eqn:diffusion},
	\begin{equation}
		\begin{split}
			\mat{x}^\ii{a}_\ii{t}
			&=\sqrt{\bar{\alpha}}_t\mat{x}^\ii{a}_0 + \sqrt{1-\bar{\alpha}_\ii{t}}\epsilon^\ii{a}\\
			&=\sqrt{\bar{\alpha}}_t\mat{x}_0 + \sqrt{1-\bar{\alpha}_\ii{t}}\epsilon^\ii{a}+\sqrt{\bar{\alpha}}_t\mat{n},
		\end{split}\label{eqn:diffusion_anomaly}
	\end{equation}
	where $\epsilon^\ii{a}$ is the noise added to $\mat{x}^\ii{a}$.
	Denote the generation process from step $\ii{t}$ by $\ii{g_t}$, by combining \cref{eqn:diffusion,eqn:diffusion_anomaly}, the error can be represented by,
	\begin{equation}
		\begin{split}
			\hat{\mat{x}}^\ii{a}-\mat{x}
			&=\ii{g_t}(\mat{x}^\ii{a}_\ii{t})-\ii{g_t}(\mat{x}_\ii{t})\\
			&=\ii{g_t}(\sqrt{\bar{\alpha}}_t\mat{x}_0 + \sqrt{1-\bar{\alpha}_\ii{t}}\epsilon^\ii{a}+\sqrt{\bar{\alpha}}_t\mat{n})-\ii{g_t}(\sqrt{\bar{\alpha}}_t\mat{x}_0 + \sqrt{1-\bar{\alpha}_\ii{t}}\epsilon)\\
			&\overset{\propto}{\sim} \sqrt{1-\bar{\alpha}}_\ii{t}(\epsilon^\ii{a}-\epsilon)+\sqrt{\bar{\alpha}_\ii{t}}\mat{n},
		\end{split}
		\label{eqn:error_analysis}
	\end{equation}
	where $\overset{\propto}{\sim}$ means approximately proportional to, which is based on a reasonable assumption that $\ii{g_t}$ is smooth enough.
	In the following, we make our efforts to reduce the errors in \cref{eqn:error_analysis} for a better reconstruction quality.

	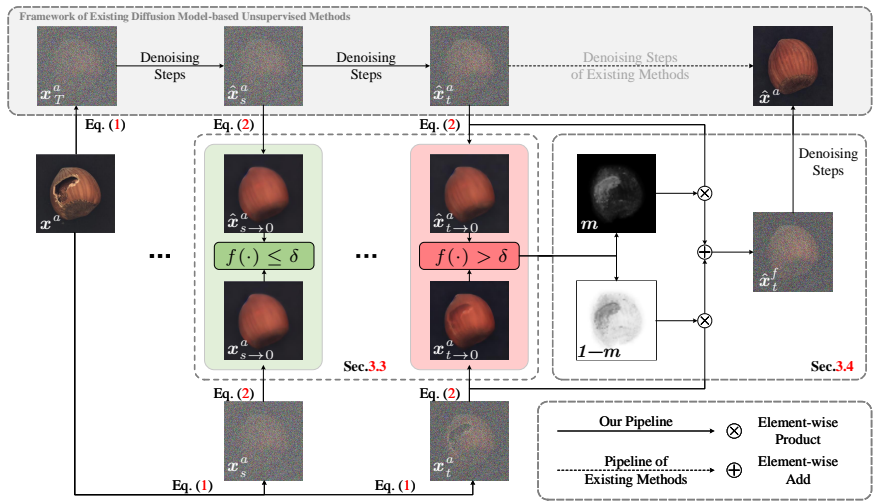
\begin{figure}[t]
		\centering
		\begin{overpic}[percent,width=.95\linewidth]{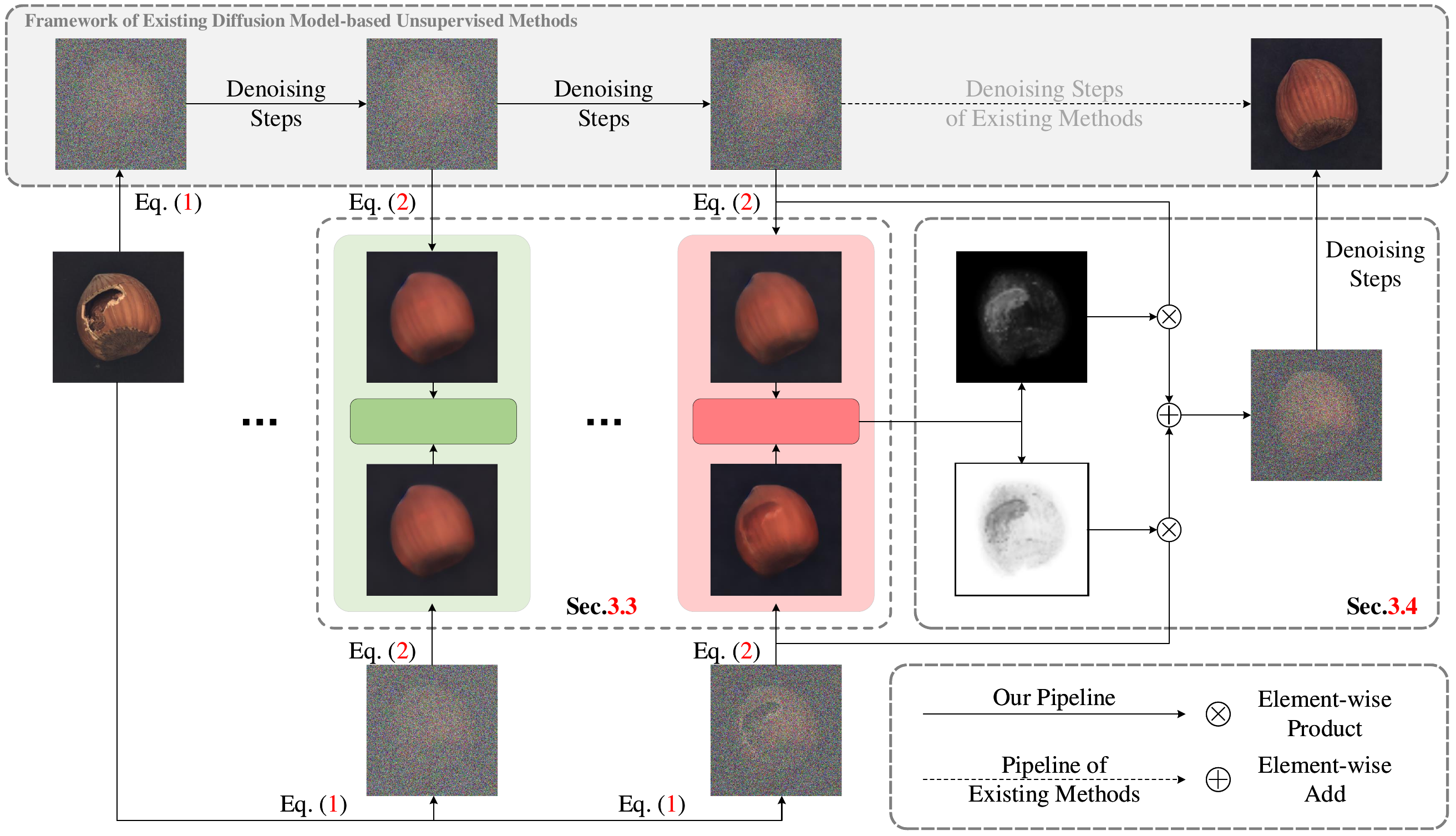}
			\put(4, 46.5){\scriptsize{$\textcolor{white}{ \mat{x}^\ii{a}_\ii{T} }$}}
			\put(4, 31.5){\scriptsize{$\textcolor{white}{ \mat{x}^\ii{a} }$}}
			
			\put(25.5,46.5){\scriptsize{$\textcolor{white}{ \hat{\mat{x}}^\ii{a}_\ii{s} }$}}
			\put(25.5,31.8){\scriptsize{$\textcolor{white}{ \hat{\mat{x}}^\ii{a}_{\ii{s}\to0} }$}}
			\put(25.5,17.2){\scriptsize{$\textcolor{white}{ \mat{x}^\ii{a}_{\ii{s}\to0} }$}}
			\put(25.5,3.7){\scriptsize{$\textcolor{white}{ \mat{x}^\ii{a}_\ii{s} }$}}
			
			\put(49.1,46.5){\scriptsize{$\textcolor{white}{ \hat{\mat{x}}^\ii{a}_\ii{t} }$}}
			\put(49.1,31.8){\scriptsize{$\textcolor{white}{ \hat{\mat{x}}^\ii{a}_{\ii{t}\to0} }$}}
			\put(49.1,17.2){\scriptsize{$\textcolor{white}{ \mat{x}^\ii{a}_{\ii{t}\to0} }$}}
			\put(49.1,3.7){\scriptsize{$\textcolor{white}{ \mat{x}^\ii{a}_\ii{t} }$}}
			
			\put(66,31.5){\scriptsize{$\textcolor{white}{ \mat{m} }$}}
			\put(66,16.8){\scriptsize{$\textcolor{black}{ \mat{\!1\!\!-\!\!m} }$}}
			
			\put(86.5,46){\scriptsize{$\textcolor{white}{ \hat{\mat{x}}^\ii{a} }$}}
			\put(86.5,25.0){\scriptsize{$\textcolor{white}{ \hat{\mat{x}}^\ii{f}_\ii{t} }$}}
			
			\put(25.5,27.7){\scriptsize{$\textcolor{black}{ \ii{f}(\cdot) \leq \delta }$}}
			\put(49.5,27.7){\scriptsize{$\textcolor{black}{ \ii{f}(\cdot) > \delta }$}}
			
		\end{overpic}
		\caption{The reconstruction pipeline of the proposed GLAD, including the Adaptive Denoising Steps (\cref{sec:method_adaptive_denoising_steps}) and the Spatial-Adaptive Feature Fusion Scheme (\cref{sec:method_spatial_adaptive_feature_fusion}).}
		\label{fig:overall}
	\end{figure}
	
	\subsection{Adaptive Denoising Steps}
	\label{sec:method_adaptive_denoising_steps}
	
	In existing methods, since the noise is always assumed to follow a standard Gaussian distribution, the error between $\epsilon^\ii{a}$ and $\epsilon$ are ignored, and then only $\sqrt{\bar{\alpha}}_\ii{t}\mat{n}$ leaves in \cref{eqn:error_analysis}.
	Considering the requirement $\|\hat{\mat{x}}^\ii{a}-\mat{x}\|_\infty<\tau$, for larger $\mat{n}$, a smaller $\sqrt{\bar{\alpha}}_\ii{t}$ is desired.
	Since $\bar{\alpha}_\ii{t}$ is negatively correlated with $\ii{t}$, a smaller $\sqrt{\bar{\alpha}}_\ii{t}$ means a larger $\ii{t}$, which provides a formulaic explanation for our motivation to set a proper denoising step for each sample.
	
	With the above analysis, an intuitive way to determine the proper step is by evaluating the value of $\mat{n}$, which, however, is a concept we introduced and is unavailable in practice.
	Actually, in the generation process, $\mat{n}$ is reflected in $\epsilon^\ii{a}_\theta=\epsilon_\theta(\mat{x}^\ii{a}_\ii{t}, t)$, which can be used for comparison.
	
	As shown in the left part of \cref{fig:overall}, starting from the input sample $\mat{x}^\ii{a}$, we add sufficient noises (\eg, $\ii{T}$ steps) and obtain $\mat{x}^\ii{a}_\ii{T}$.
	Then, in each step, we compare the following two features $\mat{x}^\ii{a}_\ii{t}$ and $\hat{\mat{x}}^\ii{a}_\ii{t}$.
	Since $\hat{\mat{x}}^\ii{a}_\ii{t}$ is generated from $\mat{x}^\ii{a}_\ii{T}$, it follows the pipeline of existing diffusion model-based methods and tends to be a normal sample.
	On the contrary, $\mat{x}^\ii{a}_\ii{t}$ is obtained by \textit{directly} adding certain noises (\ie, $\ii{t}$ steps) to $\mat{x}^\ii{a}$, and the anomalies will be preserved to some extent.
	In practice, considering that both $\hat{\mat{x}}^\ii{a}_\ii{t}$ and $\mat{x}^\ii{a}_\ii{t}$ are with noises, and the noises may deviate from each other due to the denoising procedure, we propose to convert them to the noise-free version via \cref{eqn:step_visualization}, \ie, $\hat{\mat{x}}^\ii{a}_{\ii{t}\to0}$ and $\mat{x}^\ii{a}_{\ii{t}\to0}$. And the difference between $\hat{\mat{x}}^\ii{a}_{\ii{t}\to0}$ and $\mat{x}^\ii{a}_{\ii{t}\to0}$ is measured by the anomaly score, which is calculated according to \cref{sec:method_anomaly scoring and map construction}.
	As shown in \cref{fig:overall}, if the difference is smaller than a threshold $\delta$, then we continue the denoising steps (as shown in the green part of \cref{fig:overall}) for further comparison.
	Otherwise, if the difference appears (\ie, $>\delta$) at the $\ii{t}$-th step (see the red part of \cref{fig:overall}), we can take $\mat{x}^\ii{a}_\ii{t+n}$ as the starting point to take $\ii{t+n}$ step of denoising, where $\ii{n}$ is a small number to preserve some redundancy.
	In this way, the details of normal regions are best preserved, and the anomalies can be reconstructed.
	
	\subsection{Spatial-Adaptive Feature Fusion}
	\label{sec:method_spatial_adaptive_feature_fusion}
	
	However, setting the redundant denoising step ($\ii{t+n}$) for the whole image is sub-optimal.
	As analyzed in \cref{sec:method_analysis,eqn:error_analysis}, we need only to set a larger step for the abnormal regions.
	Therefore, we argue that for normal regions, we can safely reduce the denoising step to $\ii{t}$ while keeping a large denoising step for the potential abnormal regions.
	As shown in \cref{fig:overall}, we have already performed the denoising steps from $\ii{T}$, which can be reused in this procedure.
	Specifically, we can derive a mask $\mat{m}$, which means the possibility for the pixels to be part of the anomalies.
	Then, we can combine the two features with the mask, \ie,
	\begin{equation}
		\hat{\mat{x}}^\ii{f}_\ii{t}=\mat{m}\cdot\hat{\mat{x}}^\ii{a}_\ii{t} + (1-\mat{m})\cdot\mat{x}^\ii{a}_\ii{t}.
		\label{eqn:adaptive_fusion}
	\end{equation}
	Note that the deviated noise problem still exists in \cref{eqn:adaptive_fusion}, and we follow the strategy in \cref{sec:method_adaptive_denoising_steps} to add in the noise-free version and add the same noise $\epsilon$ following \cref{eqn:diffusion} for a consistent noise, \ie,
	\begin{equation}
		\begin{split}
			\hat{\mat{x}}^\ii{f}_\ii{t}    &=\sqrt{\bar{\alpha}}_\ii{t}\hat{\mat{x}}^\ii{f}_{\ii{t}\to0}+\sqrt{1-\bar{\alpha}_\ii{t}}\epsilon\\
			&=\sqrt{\bar{\alpha}}_\ii{t}(\mat{m}\cdot\hat{\mat{x}}^\ii{a}_{\ii{t}\to0} + (1-\mat{m})\cdot\mat{x}^\ii{a}_{\ii{t}\to0})+\sqrt{1-\bar{\alpha}_\ii{t}}\epsilon.
		\end{split}\label{eqn:adaptive_fusion_noise_free}
	\end{equation}
	For a clear illustration, \cref{fig:overall} is consistent with \cref{eqn:adaptive_fusion}.
	
	\subsection{Anomaly-oriented Training Paradigm}
	\label{sec:method_anomaly_oriented_training_paradigm}
	
	With the modifications in \cref{sec:method_adaptive_denoising_steps,sec:method_spatial_adaptive_feature_fusion}, we have modulated the reconstruction process according to the properties of the anomaly detection task.
	However, as we can recall from \cref{eqn:error_analysis}, there still exists an incompatibility remaining unsolved.
	Particularly, a better reconstruction quality implies a lower value of \cref{eqn:error_analysis}, \ie,
	\begin{equation}
		\hat{\mat{x}}^\ii{a}-\mat{x}\overset{\propto}{\sim}\sqrt{1-\bar{\alpha}}_\ii{t}(\epsilon^\ii{a}-\epsilon)+\sqrt{\bar{\alpha}_\ii{t}}\mat{n}\to0.
		\label{eqn:error_to_0}
	\end{equation}
	In \cref{eqn:error_to_0}, $\bar{\alpha}_\ii{t}$, $\epsilon$, and $\mat{n}$ are manually set values or inherent concepts.
	The only value to be estimated by the model is $\epsilon^\ii{a}$.
	By rewriting \cref{eqn:error_to_0}, it should follow,
	\begin{equation}
		\epsilon^\ii{a}\to\epsilon-\tfrac{\sqrt{\bar{\alpha}_\ii{t}}}{\sqrt{1-\bar{\alpha}_\ii{t}}}\mat{n}.
		\label{eqn:error_of_predicted_noise}
	\end{equation}
	Following the setting of diffusion models, $\epsilon\sim\mathcal{N}(\bb{0}, \bb{I})$ is a random noise following standard Gaussian distribution.
	Once $\mat{n}$ is non-zero (\ie, anomalies exist), we can draw a conclusion that $\epsilon^\ii{a}$ deviates from the standard Gaussian distribution.
	
	For a typical diffusion model ($\epsilon_\theta$) trained with normal samples only, the noises in all steps follow the standard Gaussian distribution.
	In other words, it is beyond the scope of $\epsilon_\theta$ to predict such an $\epsilon^\ii{a}$.
	Therefore, we propose introducing anomalies during training, which enables the diffusion models to break through the limitation of the standard Gaussian distribution and fit \cref{eqn:error_of_predicted_noise}.
	On the basis of \cref{eqn:error_of_predicted_noise}, the learning objective can be formulated by,
	\begin{equation}
		\begin{split}
			\ii{L_{ATP}}&=\mathbb{E}_{(\mat{x},\mat{x}^\ii{a})\sim\ii{p_{data}},\epsilon\sim\mathcal{N}(\bb{0}, \bb{I}),\ii{t}}[\|(\epsilon-\tfrac{\sqrt{\bar{\alpha}_\ii{t}}}{\sqrt{1-\bar{\alpha}_\ii{t}}}\mat{n})-\epsilon^\ii{a}\|_2]\\
			&=\mathbb{E}_{(\mat{x},\mat{x}^\ii{a})\sim\ii{p_{data}},\epsilon\sim\mathcal{N}(\bb{0}, \bb{I}),\ii{t}}[\|(\epsilon-\tfrac{\sqrt{\bar{\alpha}_\ii{t}}}{\sqrt{1-\bar{\alpha}_\ii{t}}}(\mat{x}^\ii{a}-\mat{x}))-\epsilon_\theta(\mat{x}^\ii{a}_\ii{t},\ii{t})\|_2].
		\end{split}
		\label{eqn:loss}
	\end{equation}
	One can see that \cref{eqn:loss} places even higher demands on the datasets than supervised anomaly detection since the corresponding normal sample is desired, which is difficult or even infeasible to prepare.
	To remedy the dilemma, we follow MemSeg~\cite{MemSeg19} to synthesize abnormal samples with normal ones, which enables the training to proceed in an unsupervised manner.
	
	It is obvious that the \cref{eqn:loss} is a more general form. For the normal region, the formula can degenerate to the original diffusion loss, while for the abnormal region, the formula forces the model to predict non-Gaussian noise to result in the corresponding normal region.
	
	\subsection{Anomaly Scoring and Map Construction}
	\label{sec:method_anomaly scoring and map construction}
	Following AprilGAN~\cite{APRILGAN18}, we regard reconstructed images as reference images to compare with test images to construct anomaly maps. A pre-trained model (DINO~\cite{DINO39}) is utilized to extract the multi-layer features $\ii{F_t}$ of test images and $\ii{F_r}$ of reconstructed images, respectively.
	The anomaly map $\ii{M_l}\in \mathbb{R}^{\ii{u} \times \ii{v}}$ of layer $\ii{l}$ are calculated based on the cosine similarity between layer $\ii{l}$ features $\ii{F_t^l} \in \mathbb{R}^{\ii{c} \times \ii{u} \times \ii{v}} $ and $\ii{F_r^l}\in \mathbb{R}^{\ii{c} \times \ii{u} \times \ii{v}}$, \ie,
	\begin{equation}
		\ii{M_l}^{(\ii{i}, \ii{j})}(\ii{F_t^l},\ii{F_r^l})=\min(1 - \langle \ii{F_t^l}^{(\ii{i}, \ii{j})}, \ii{F_r^l} \rangle),
	\end{equation}
	where $(\ii{i}, \ii{j})$ is the coordinate, $ \langle \ii{x},\mat{y} \rangle$ calculates the cosine similarity between $\ii{x}$ and all elements of $\mat{y}$, and $\min(\cdot)$ returns the minimal value.
	Finally, the anomaly maps of different layers are added as the anomaly map $\ii{M}\in \mathbb{R}^{\ii{u} \times \ii{v}}$, \ie,
	\begin{equation}
		\ii{M}={\sum}_l \ii{M_l}(\ii{F_t^l}, \ii{F_r^l}),
	\end{equation}
	and the anomaly score of the whole image is the average of the top $\ii{K}$ maximum values of $\ii{M}$.
	
	\section{Experiments}
	In this section, we first introduce the experiment setup and details of its implementation. Then, we compare our method with state-of-the-art (SOTA) unsupervised anomaly detection methods to show the superiority of our method. In addition, To evaluate the effectiveness of our method, we conduct ablation studies on the individual components of our method. 
	
	\subsection{Experiments Setup}
	\noindent\textbf{Datasets.}
	We conduct experiments on two challenging datasets, MVTec-AD~\cite{MVTec-AD28} MPDD~\cite{MPDD29}, VisA~\cite{visa40}, and PCB-Bank datasets, to evaluate the effectiveness of our method.
	
	\textit{MVTec-AD.} MVTec-AD contains 10 objects and 5 texture classes of industrial anomalous samples with mask annotations. There are 3,629 images for training/validation and 1,725 images for testing.
	
	\textit{MPDD.} MPDD contains 6 classes of metal parts, comprising 888 normal samples for the training set and 458 samples either normal or anomalous in the test set. Because of the variable spatial orientation, position, and distance of multiple objects concerning the camera at different light intensities and with a non-homogeneous background, this dataset is a more challenging dataset.
	
	\textit{VisA.} VisA is twice the size of MVTec, comprising 9,621 normal and 1,200 anomalous high-resolution images. This dataset exhibits objects of complex structures placed in sporadic locations as well as multiple objects in one image. Anomalies include scratches, dents, color spots, cracks, and structural defects.
	
	\textit{PCB-Bank.} PCB-Bank is a printed circuit board dataset we integrated, including 7 different categories. Class PCB1$\sim$PCB4 are from VisA~\cite{visa40} dataset, Class PCB5 from RealIAD~\cite{realiad} dataset and Class PCB6$\sim$PCB7 from VISION~\cite{vision} dataset. There are 4214 normal samples for the training set and 2253 samples that are either normal or anomalous in the test set. The samples of the dataset have different clarity, resolution, and shooting angle. Abnormal types mainly include scratches, structural defects, bends, and \etc.
	
	\vspace{0.5em}
	\noindent\textbf{Evaluation Metrics.}
	Following prior works, we report the two most widely used metrics, image-level Area Under the Receiver Operating Curve (I-AUROC) for anomaly detection and pixel-level Area Under the Receiver Operating Curve (P-AUROC) for anomaly localization. In addition, to further verify the superiority of the proposed method, we report on Average Precision (AP) and F1-score-max (F1-max) in both anomaly detection and anomaly localization, where the prefix I- and prefix P- stand for anomaly detection and anomaly localization, respectively. Also, Per-Region-Overlap (PRO) is used in anomaly localization.
	
	\subsection{Implementation Details}
	We use the pre-trained latent diffusion model (LDM)~\cite{SD30} and fine-turn the UNet to adapt data for reconstruction. DINO~\cite{DINO39} with ViT-B/8 architectures is utilized as a feature extraction model. We only fine-tune the VAE and DINO for the VisA and PCB-Bank datasets because of the larger differences between the datasets and the pre-training datasets. To be consistent with the pre-trained VAE, images are resized to resolutions of 512 $\times$ 512. We also report the results of 256 $\times$ 256 resolutions in \cref{table_7results}. Features of layers 3, 6, 9, and 12 are used for anomaly map construction. The UNet is trained with 4,000 iterations. The batch size is 32, and the learning rate is set as $5 \times 10^{-6}$. $T$ is generally set as 750 for the MVTec-AD dataset, 500 for the MPDD dataset, and 450 for the VisA and PCB-Bank datasets. A minimum step $t_{min}$ is set as 350 for the MVTec-AD dataset and MPDD dataset, and 200 for the VisA and PCB-Bank dataset to avoid missing anomalies. A Gaussian filter with $\sigma = 6$ is used to smooth the anomaly localization score, and the average of the top 250 maximum values of $\ii{M}$ is considered as the anomaly score. In \cref{sec:method_adaptive_denoising_steps}, we use the feature of layer 12 and top 10 maximum values to calculate difference. More details of fine-tuning DINO and parameter for multi-category settings are included in the supplementary material.
	
	\subsection{Comparison with State-of-the-art Methods}
	\begin{table}
		\caption{Comparison with SOTA methods on MVTec-AD dataset. I-AUROC and P-AUROC are displayed in each entry. The best results among all methods are shown in bold, and the underlined results denote the best results among reconstruction-based methods.}
		\label{table_mvtec}
		\centering
		\resizebox{1.0\columnwidth}{!}{
			\begin{tabular}{cccc|cccccc}
				\toprule
				\multirow{2}{*}{ Category } & \multicolumn{3}{c|}{ Embedding-based methods } & \multicolumn{6}{c}{ Reconstruction-based methods } \\
				\cline{2-10} 
				& PatchCore~\cite{PatchCore24} & RD4AD~\cite{RD4AD14} & SimpleNet~\cite{Simplenet23}  & DRAEM~\cite{DRAEM7} & OCR-GAN~\cite{OCR-GAN5} &Lu~\etal~\cite{26} & DiffAD~\cite{DiffAD9} & DDAD~\cite{DDAD8} & Ours \\
				\hline
				Carpet     & 98.7/\textbf{99.0} & 98.9/98.8  & \textbf{99.7}/98.2  & 97.0/95.5 & \underline{99.4}/- & -/97.7 &98.3/98.1 &99.3/\underline{98.7} & 99.0/98.5\\
				Grid       & 98.2/98.7 & \textbf{100}/97.0  & 99.7/98.8  & 99.9/\underline{\textbf{99.7}} & 99.6/- & -/95.6  &\underline{\textbf{100}}/\underline{\textbf{99.7}} &\underline{\textbf{100}}/99.4 & \underline{\textbf{100}}/99.6\\
				Leather    & \textbf{100}/99.3  & \textbf{100}/98.6   & \textbf{100}/99.2   & \underline{\textbf{100}}/98.5  & 97.1/- & -/97.5  &\underline{\textbf{100}}/99.1 &\underline{\textbf{100}}/99.4 & \underline{\textbf{100}}/\underline{\textbf{99.8}}\\
				Tile       & 98.7/95.6 & 99.3/98.9  & 99.8/97.0  & 99.6/99.2 & 95.5/- & -/98.9 &\underline{\textbf{100}}/\underline{\textbf{99.4}} &\underline{\textbf{100}}/98.2 & \underline{\textbf{100}}/98.7\\
				Wood       & 99.2/95.0 & 99.2/\textbf{99.3}  & \textbf{100}/94.5   & 99.1/96.4 & 95.7/- & -/\underline{99.1} &\underline{\textbf{100}}/96.7 &\underline{\textbf{100}}/95.0 & 99.4/98.4\\
				\hline
				Bottle     & \textbf{100}/98.6  & \textbf{100}/99.0   & \textbf{100}/98.0  & 99.2/\underline{\textbf{99.1}} & 99.6/- & -/97.3 &\underline{\textbf{100}}/98.8  &\underline{\textbf{100}}/98.7 & \underline{\textbf{100}}/98.9\\
				Cable      & 99.5/98.4 & 95.0/99.4  & \textbf{99.9}/97.6 & 91.8/94.7 & 99.1/- & -/\underline{\textbf{99.5}} &94.6/96.8 &99.4/98.1 & \underline{\textbf{99.9}}/98.1\\
				Capsule    & 98.1/98.8 & 96.3/97.3  & 97.7/\textbf{98.9} & 98.5/94.3 & 96.2/- & -/96.8 &97.5/98.2 &99.4/95.7 & \underline{\textbf{99.5}}/\underline{98.5}\\
				Hazelnut   & \textbf{100}/98.7  & 99.9/98.2   & \textbf{100}/97.9  & \underline{\textbf{100}}/\underline{\textbf{99.7}}  & 98.5/- & -/92.5  &\underline{\textbf{100}}/99.4  &\underline{\textbf{100}}/98.4 & \underline{\textbf{100}}/99.5\\
				Metal nut  & \textbf{100}/98.4  & \textbf{100}/\textbf{99.6}  & \textbf{100}/98.8  & 98.7/\underline{99.5} & 99.5/- & -/99.0  &\underline{\textbf{100}}/99.4  &\underline{\textbf{100}}/99.0 & \underline{\textbf{100}}/98.8\\
				Pill       & 96.6/97.4 & \textbf{99.6}/95.7  & 99.0/\textbf{98.6} & 98.9/97.6 & 98.3/- & -/92.1 &97.7/97.7 &\underline{\textbf{100}}/99.1 & 98.1/\underline{97.9}\\
				Screw      & 98.1/\textbf{99.4} & 97.0/99.1  & 98.2/99.3 & 93.9/97.6 & \underline{\textbf{100}}/-  & -/98.6 &97.2/99.0 &99.0/\underline{99.3} & 96.9/99.1\\
				Toothbrush & \textbf{100}/98.7  & 99.5/93.0  & 99.7/98.5 & \underline{\textbf{100}}/98.1  & 98.7/- & -/93.1  &\underline{\textbf{100}}/99.2  &\underline{\textbf{100}}/98.7 & \underline{\textbf{100}}/\underline{\textbf{99.4}}\\
				Transistor & \textbf{100}/96.3  & 96.7/95.4  & \textbf{100}/\textbf{97.6}  & 93.1/90.9 & \underline{98.3}/- & -/94.5 &96.1/93.7 &\underline{\textbf{100}}/95.3 & 98.3/\underline{96.2}\\
				Zipper     & 99.4/98.8 & 98.5/98.2  & 99.9/98.9 & \underline{\textbf{100}}/98.8  & 99.0/- & -/97.6  &\underline{\textbf{100}}/\underline{\textbf{99.0}}  &\underline{\textbf{100}}/98.2 & 98.5/97.9\\
				\hline
				Average    & 99.1/98.1 & 98.5/97.8 & 99.6/98.1  & 98.0/97.3 & 98.3/- & -/96.7  & 98.7/98.3 &\underline{\textbf{99.8}}/98.1 & 99.3/\underline{\textbf{98.6}}\\
				\bottomrule
			\end{tabular}
		}
	\end{table}
	The comparisons between state-of-the-art (SOTA) methods and our method are shown in \cref{table_mvtec} for the MVTec-AD dataset. We compare our method with embedding-based methods (PatchCore~\cite{PatchCore24}, RD4AD~\cite{RD4AD14} and SimpleNet~\cite{Simplenet23}), and reconstruction-based methods (DRAEM~\cite{DRAEM7}, OCR-GAN~\cite{OCR-GAN5}, Lu~\etal~\cite{26}, DiffAD~\cite{DiffAD9} and DDAD~\cite{DDAD8}). Lu~\etal, DiffAD, and DDAD are advanced diffusion-based methods. For image-level anomaly detection tasks, our method achieves the highest I-AUROC on 9 out of 15 classes. Although the I-AUROC of our method is slightly lower than DDAD's, our method is superior to DDAD on I-AP and I-F1-max (99.7/98.4 VS 99.5/97.9) in \cref{table_7results}. For pixel-level anomaly localization tasks, our method outperforms the SOTA in terms of P-AUROC among all types of methods. As shown in \cref{table_7results}, our method exceeds reconstruction-based SOTA (DDAD) by 11.9↑/9.6↑/3.0↑ in P-AP/P-F1-max/PRO.
	
	\begin{table}
		\caption{Comparison with SOTA on MPDD dataset. I-AUROC and P-AUROC are displayed in each entry. The best results among all methods are shown in bold, and the underlined results denote the best results among reconstruction-based methods.}
		\label{table_mpdd}
		\centering
		\resizebox{1.0\columnwidth}{!}{
			\begin{tabular}{cccc|cccc}
				\toprule
				\multirow{2}{*}{ Category } & \multicolumn{3}{c|}{ Embedding-based methods } & \multicolumn{4}{c}{ Reconstruction-based methods } \\
				\cline{2-8}
				&PatchCore~\cite{PatchCore24} &RD4AD~\cite{RD4AD14} &SimpleNet~\cite{Simplenet23} &DRAEM~\cite{DRAEM7} &OCR-GAN~\cite{OCR-GAN5} &DDAD~\cite{DDAD8}  &Ours\\
				\hline
				Bracket Black   &85.3/97.6  &90.2/98.0  &85.1/96.0 &91.8/98.2 &99.9/- &\underline{\textbf{98.7}}/96.7   &98.0/\underline{\textbf{99.4}} \\
				Bracket Brown   &92.5/\textbf{98.1}  &94.2/97.0  &\textbf{98.3}/94.4 &90.3/63.7 &89.4/- &92.7/97.2   &\underline{90.7}/\underline{97.5}\\
				Bracket White   &92.3/99.7  &90.1/99.3  &98.0/96.7 &88.8/98.9 &88.1/- &96.6/91.8   &\underline{\textbf{98.3}}/\underline{\textbf{99.7}}\\
				Connector       &100/99.4  &99.5/99.4  &\textbf{100}/\textbf{99.5} &\underline{\textbf{100}}/91.2 &100/- &96.2/98.6   &\underline{\textbf{100}}/\underline{98.2}\\
				Metal Plate     &\textbf{100}/98.8   &\textbf{100}/99.1   &\textbf{100}/98.5  &\underline{\textbf{100}}/96.6  &100/- &100/98.1   &99.9/\underline{\textbf{99.4}}\\
				Tubes           &77.4/97.2  &97.6/99.1   &97.9/\textbf{99.2} &94.7/95.9 &98.1/- &99.2/99.0   &\underline{\textbf{98.1}}/\underline{97.8}\\
				\hline
				Average         &91.3/98.5  &95.3/\textbf{98.7}  &96.6/97.4 &94.3/90.7 &95.9/- &97.2/96.9   &\underline{\textbf{97.5}}/\underline{\textbf{98.7}} \\
				\bottomrule
			\end{tabular}
		}
	\end{table}          
	
	We also conduct experiments on the MPDD dataset to further prove the superiority of the proposed method in \cref{table_mpdd}. Our method outperforms the SOTA in terms of both I-AUROC and P-AUROC among all types of methods. In addition, our method surpasses the reconstruction-based SOTA by 1.8↑/10.5↑/5.7↑/5.7↑ on P-AUROC/P-AP/P-F1-max/PRO as shown in \cref{table_7results}.  
	
	\cref{table_visa,table_pcbbank} show the results of VisA and PCB-Bank datasets. Our method also achieved SOTA performance both on anomaly detection and localization tasks (99.5/98.6 on the VisA dataset and 98.7/99.3 on the PCB-Bank dataset on I-AUROC and P-AUROC).
	\begin{table}
		\caption{Comparison with SOTA methods on VisA dataset. I-AUROC and P-AUROC are displayed in each entry. The best results among all methods are shown in bold, and the underlined results denote the best results among reconstruction-based methods.}
		\label{table_visa}
		\centering
		\resizebox{1.0\columnwidth}{!}{
			\begin{tabular}{cccc|cccc}
				\toprule
				\multirow{2}{*}{ Category } & \multicolumn{3}{c|}{ Embedding-based methods } & \multicolumn{4}{c}{ Reconstruction-based methods } \\
				\cline{2-8} 
				& PatchCore~\cite{PatchCore24} & RD4AD~\cite{RD4AD14} & SimpleNet~\cite{Simplenet23}  & DRAEM~\cite{DRAEM7} & OCR-GAN~\cite{OCR-GAN5} &DDAD~\cite{DDAD8} & Ours \\
				\hline
				Candle     & 98.7/\textbf{99.2} & 96.2/98.9  & 96.9/98.6  & 89.6/91.0 & 98.9/- &\underline{\textbf{99.9}}/\underline{98.7} & \underline{\textbf{99.9}}/94.8\\
				Capsules   & 68.8/96.5 & 91.8/99.4  & 89.5/99.2  & 89.2/99.0 & 98.8/- &\underline{\textbf{100}}/99.5  & 99.1/\underline{\textbf{99.6}}\\
				Cashew     & 97.7/99.2 & \textbf{98.7}/94.4  & 94.8/\textbf{99.0}  & 88.3/85.0 & 97.4/- &94.5/\underline{97.4} & \underline{98.4}/97.0\\
				Chewinggum & 99.1/98.9  & 99.3/97.6  & \textbf{100}/98.5   & 96.4/97.7 & 99.4/- &98.1/96.5 & \underline{99.6}/\underline{\textbf{99.1}}\\
				Fryum      & 91.6/95.9 & 96.9/96.4  & 96.6/94.5  & 94.7/82.5 & 96.3/- &99.0/\underline{\textbf{96.9}} & \underline{\textbf{99.4}}/\underline{\textbf{96.9}}\\
				Macaroni1  & 90.1/98.5 & 98.7/99.3  & 97.6/99.6  & 93.9/99.4 & 97.2/- &99.2/98.7 & \underline{\textbf{99.9}}/\underline{\textbf{99.8}}\\
				Macaroni2  & 63.4/93.5 & 91.4/99.1  & 83.4/96.4  & 88.3/99.7 & 95.1/- &\underline{\textbf{99.2}}/98.2 & 98.9/\underline{\textbf{99.8}}\\
				Pcb1       & 96.0/\textbf{99.8}& 96.7/99.6  & 99.2/\textbf{99.8}  & 84.7/98.4 & 96.1/- &\underline{\textbf{100}}/93.4  & 99.6/\underline{99.6}\\
				Pcb2       & 95.1/98.4 & 97.2/98.3  & 99.2/\textbf{98.8}  & 96.2/94.0 & 98.3/- &99.7/97.4 & \underline{\textbf{100}}/ \underline{98.6}\\
				Pcb3       & 93.0/98.9 & 96.5\textbf{/99.3}  & 98.6/99.2  & 97.4/94.3 & 98.1/- &97.2/96.3 & \underline{\textbf{99.9}}/\underline{98.9}\\
				Pcb4       & 99.5/98.3 & 99.4/98.2   & 98.9/98.6  & 98.9/97.6 & 99.7/- &\underline{\textbf{100}}/98.5  & 99.9/\underline{\textbf{99.5}}\\
				Pipe fryum & 99.0/99.3 & 99.6/99.1  & 99.2/99.3  & 94.7/65.8 & 99.7/- &\underline{\textbf{100}}/\underline{\textbf{99.5}}  & 98.9/\underline{99.4}\\
				\hline
				Average    & 91.0/98.1 & 96.9/98.3 & 96.2/98.5  & 92.4/92.0 & 97.9/- & 98.9/97.6 & \underline{\textbf{99.5}}/\underline{\textbf{98.6}}\\
				\bottomrule
			\end{tabular}
		}
	\end{table}
	
	\begin{table}
		\caption{Comparison with SOTA methods on PCB-Bank dataset. I-AUROC and P-AUROC are displayed in each entry. The best results among all methods are shown in bold, and the underlined results denote the best results among reconstruction-based methods.}
		\label{table_pcbbank}
		\centering
		\resizebox{1.0\columnwidth}{!}{
			\begin{tabular}{cccc|cccc}
				\toprule
				\multirow{2}{*}{ Category } & \multicolumn{3}{c|}{ Embedding-based methods } & \multicolumn{4}{c}{ Reconstruction-based methods } \\
				\cline{2-8} 
				& PatchCore~\cite{PatchCore24} & RD4AD~\cite{RD4AD14} & SimpleNet~\cite{Simplenet23}  & DRAEM~\cite{DRAEM7} & OCR-GAN~\cite{OCR-GAN5} &DDAD~\cite{DDAD8} & Ours \\
				\hline
				Pcb1       & 96.0/\textbf{99.8}& 96.7/99.6  & 99.2/\textbf{99.8}  & 84.7/98.4 & 96.1/- &\underline{\textbf{100}}/93.4  & 99.6/\underline{99.6}\\
				Pcb2       & 95.1/98.4 & 97.2/98.3  & 99.2/\textbf{98.8}  & 96.2/94.0 & 98.3/- &99.7/97.4 & \underline{\textbf{100}}/ \underline{98.6}\\
				Pcb3       & 93.0/98.9 & 96.5\textbf{/99.3}  & 98.6/99.2  & 97.4/94.3 & 98.1/- &97.2/96.3 & \underline{\textbf{99.9}}/\underline{98.9}\\
				Pcb4       & 99.5/98.3 & 99.4/98.2   & 98.9/98.6  & 98.9/97.6 & 99.7/- &\underline{\textbf{100}}/98.5  & 99.9/\underline{\textbf{99.5}}\\
				Pcb5       & 94.6/\textbf{99.8} & 94.1/99.5  & 94.5/99.4 & 97.2/97.8 & 85.9/- & \underline{\textbf{99.7}}/96.0  & 99.6/\underline{99.1}\\
				Pcb6       & 82.2/98.9 & 89.4/98.9  & 91.7/97.5 & 72.4/94.6 & 75.1/-  & 87.8/98.5 & \underline{\textbf{92.2}}/\underline{\textbf{99.7}} \\
				Pcb7       & 93.7/98.8 & 99.0/99.6  & \textbf{100}/\textbf{99.9}  & 97.7/98.3 & 85.7/- &94.4/98.7 & \underline{99.6}/\underline{99.8}\\
				\hline
				Average    & 94.2/99.1 & 96.0/99.1 & 96.2/98.5  & 91.5/96.4 & 91.3/- & 97.4/96.5 & \underline{\textbf{98.7}}/\underline{\textbf{99.3}}\\
				\bottomrule
			\end{tabular}
		}
	\end{table}
	
	\cref{table_7results} further reports comprehensive comparisons between SOTA methods and our GLAD. I-AUROC, I-AP, and I-F1-max are shown at the first raw of each method for anomaly detection performance, while P-AUROC, P-AP, P-F1-max, and PRO are for anomaly localization performance at the second raw. Overall, our approach outperforms existing approaches on all metrics on the average of the four datasets. In particular, because large-scale abnormal regions can be reconstructed into normal regions, our method has great advantages in anomaly location results.
	
	Reconstructions and qualitative results on MVTec-AD and MPDD datasets are displayed in \cref{fig:heatmap}. More quantitative results are presented in supplementary materials. For reconstruction results, other methods usually fail to reconstruct large-scale anomalies into normal regions. On the contrary, our method can produce satisfactory reconstruction results. In (f)-(h) column of qualitative results, other methods ignore some abnormal areas and produce inaccurate locations because of failed reconstructions of abnormal areas. However, because our method can guarantee anomaly-free reconstruction, our method can accurately locate anomalies, as shown in the (i) column.  and therefore can accurately locate the abnormal area. It also explains the superior quantification results for the location of our method compared with other reconstruction-based methods in \cref{table_7results}.

	\begin{table}
		\caption{Quantitative results on MVTec-AD, MPDD, VisA and PCB-Bank datasets. Metrics are I-AUROC/I-AP/I-F1-max at first raw (for detection) and P-AUROC/P-AP/P-F1-max/PRO at second raw (for localization).}\label{tab:more_exp}
		\centering
		\label{table_7results}
		\resizebox{1.0\columnwidth}{!}{
			\begin{tabular}{ccccc|c}
				\hline
				\makebox[0.1\textwidth][c]{Dataset}   &\makebox[0.25\textwidth][c]{MVTec-AD}  & \makebox[0.25\textwidth][c]{MPDD}               &\makebox[0.25\textwidth][c]{VisA}           & \makebox[0.25\textwidth][c]{PCB-Bank}  &\makebox[0.25\textwidth][c]{Avg}     \\      
				\hline
				\multirow{2}{*}{PatchCore[1]}    &99.1/99.6/98.1              &91.3/95.1/91.3       &91.0/92.7/88.7       & 94.2/95.6/90.3       &93.9/95.8/92.1\\
				&98.1/55.9/57.6/93.4         &98.5/38.4/40.7/92.9  &98.1/38.5/40.5/88.3  & 99.1/46.0/48.5/90.8  &98.5/44.7/46.9/91.4\\
				
				\multirow{2}{*}{RD4AD[2]}         &98.7/99.5/98.0               &95.3/96.8/93.0      &96.9/97.2/93.8      & 96.0/96.2/92.6        &96.8/97.5/94.4\\
				&97.9/59.0/61.2/94.1          &98.7/44.5/46.1/95.2 &98.3/\textbf{44.6/47.2}/93.0 & 99.1/46.3/48.0/94.0   &98.5/48.9/50.8/94.2\\
				
				\multirow{2}{*}{SimpleNet[3]}    &99.6/99.6/\textbf{98.9}     &96.6/97.7/96.0      &96.2/96.9/92.6      & 97.4/98.1/94.6        &97.5/98.1/95.5\\
				&98.1/49.8/52.8/91.9          &97.4/35.6/37.5/90.4 &98.5/33.2/37.1/92.3 & 99.0/43.3/45.2/94.4   &98.3/40.5/43.2/92.3\\
				\hline
				\multirow{2}{*}{DRAEM[4]}        &98.0/99.0/96.9      &94.3/95.8/93.0       &92.4/93.4/87.9          & 91.5/91.8/88.2             &94.1/95.0/91.5\\
				&97.3/68.4/66.7/91.3 &90.7/28.3/29.8/78.0  &92.0/28.8/36.1/78.7     & 96.4/32.2/38.0/80.9        &94.1/39.4/42.7/82.2\\
				
				\multirow{2}{*}{OCR-GAN[5]}      &98.3/98.1/95.0      &96.2/96.6/\textbf{97.7}       &97.9/98.7/96.4          & 91.3/91.6/88.0      &95.9/96.3/94.3\\
				&-/-/-/-             &-/-/-/-              &-/-/-/-                 & -/-/-/-             &/-/-/-\\
				
				\multirow{2}{*}{DDAD[6]}        &\textbf{99.8}/99.5/97.9       &97.2/95.5/95.1       &98.9/98.6/96.2          & 97.4/96.1/95.2      &98.3/97.4/96.1\\
				&98.1/59.0/59.4/92.3  &96.9/34.8/43.5/91.6  &97.6/27.9/34.6/92.7     & 96.5/28.1/33.6/91.1 &97.3/37.5/42.8/91.9\\  
				\hline                                             
				\multirow{2}{*}{Ours-256}    &99.0/99.7/98.2       &97.3/98.4/95.4       &99.3/\textbf{99.6}/97.6           &98.1/98.4/96.7        &98.4/98.8/97.0\\
				&\textbf{98.7}/63.8/63.7/95.2  &98.5/41.5/43.9/94.2  &98.3/35.8/42.4/94.1      &98.8/42.0/47.5/93.8   &98.6/44.5/48.6/94.3\\  
				
				\multirow{2}{*}{Ours-512}   &99.3/\textbf{99.7}/98.4       &\textbf{97.5/98.5}/96.0       &\textbf{99.5}/99.4/\textbf{98.3}            & \textbf{98.7/99.0/97.3}      &\textbf{98.8/99.2/97.5}\\
				&98.6\textbf{/70.9/69.0/95.3}  &\textbf{98.7/45.3/49.2/96.3}  &\textbf{98.6}/39.1/45.4/\textbf{94.3}       & \textbf{99.3/48.9/52.2/95.1} &\textbf{98.8/51.1/54.0/95.3}\\  
				\bottomrule   
			\end{tabular}
		}
	\end{table}
	
	\begin{figure}
		\centering
		\includegraphics[width=1.0\linewidth]{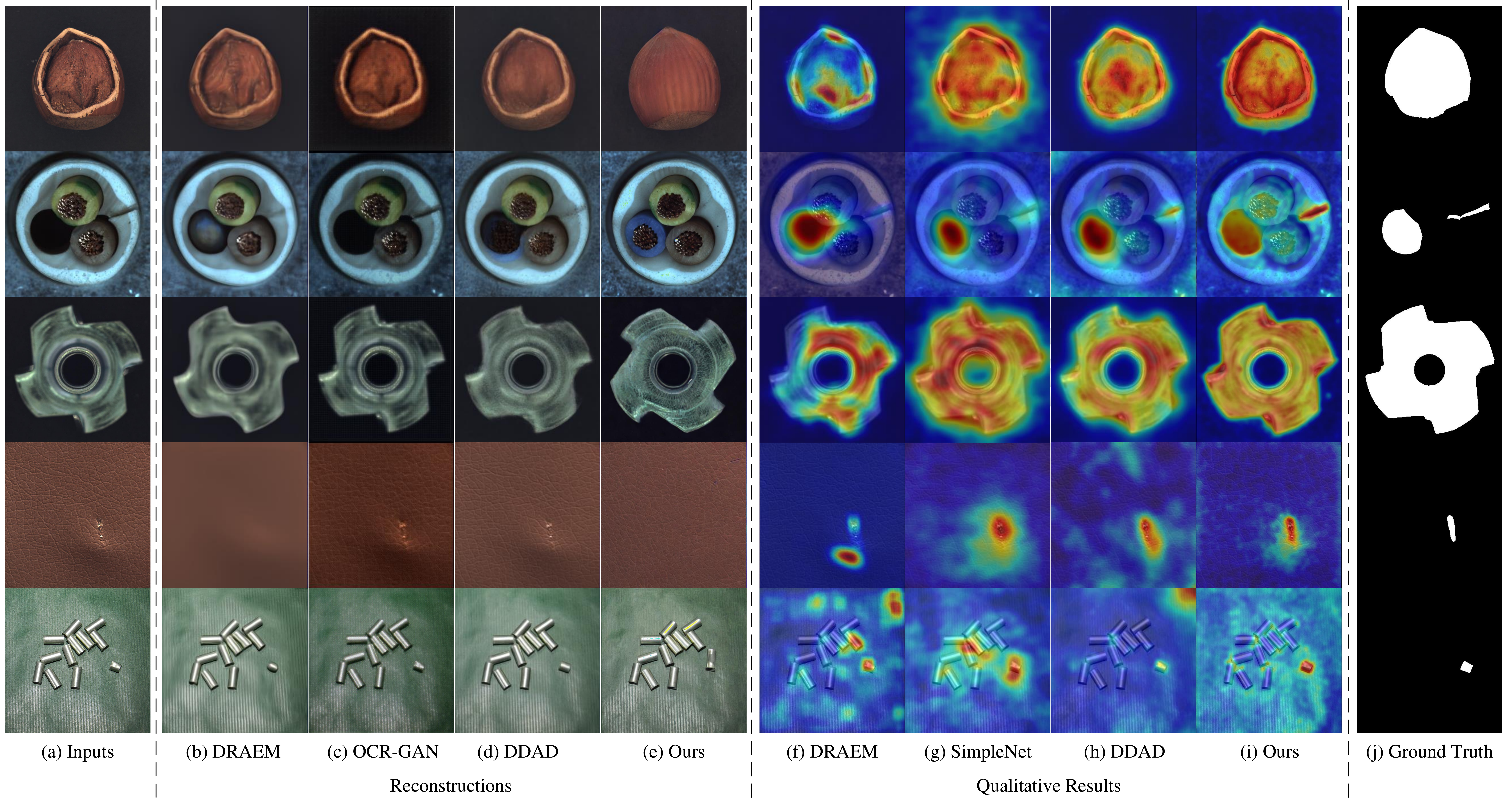}
		\caption{Reconstructions and qualitative comparisons with other methods. The first four rows display examples of the MVTec-AD dataset, and the last row is for the MPDD dataset. OCR-GAN only produces anomaly scores, and there is no anomaly map. SimpleNet is the embedding-based method.
		}
		\label{fig:heatmap}
	\end{figure}
	
	\subsection{Ablation Study}
	We propose three components, \ie, Adaptive Denoising Steps (ADS), Spatial-Adaptive Feature Fusion (SAFF) and Anomaly-oriented Training Paradigm (ATP). In this section, ablation studies are conducted to verify the effectiveness of our proposed components. A discussion of hyperparameters is presented in the supplementary materials.
	
	\subsubsection{Adaptive Denoising Steps}
	
	With the proposed ADS, DM chooses the proper noise steps and denoises from corresponding noisy samples, which ensures anomaly-free reconstruction and preserves as much information as possible about normal regions. In \cref{table3}, the improvement brought by ADS is obvious. We also compare our method with other cases which use different fixed steps in \cref{table4}. The experimental results prove the superiority of the proposed method. 
	\begin{table}
		\caption{Performance of each component on MVTec-AD dataset. The best results are shown in bold.}
		\label{table3}
		\centering
		\begin{tabular}{c|cc}
			\toprule
			Method                                             &I-AUROC  &P-AUROC \\
			\hline
			Baseline                                           &98.3     &98.0 \\
			Baseline $\ii+$ ADS                                &99.0     &98.5 \\
			Baseline $\ii+$ ATP                                &98.7     &98.5 \\
			Baseline $\ii+$ ADS $\ii+$ ATP w/o SAFF            &99.2     &98.3 \\
			Baseline $\ii+$ ADS $\ii+$ ATP with SAFF (Ours)    &\textbf{99.3}  &\textbf{98.6} \\
			\bottomrule
		\end{tabular}
	\end{table}
	
	\begin{table}
		\caption{Comparison of adaptive steps and different fixed steps on MVTec-AD dataset. The best results are shown in bold.}
		\label{table4}
		\centering
		\begin{tabular}{c|cccc}
			\toprule
			\multirow{2}{*}{ Denoising steps }  & \multicolumn{3}{c}{ fixed steps } &\multirow{2}{*}{ Adaptive steps }  \\
			\cline{2-4}
			&350 step  &550 step &750 step  &\\
			\hline
			I-AUROC   &98.8  &98.8  &98.7  &\textbf{99.3} \\
			P-AUROC   &97.6  &98.1  &98.5  &\textbf{98.6} \\
			\bottomrule
		\end{tabular}
	\end{table}
	
\begin{figure}
		\centering
		\includegraphics[width=1.0\linewidth]{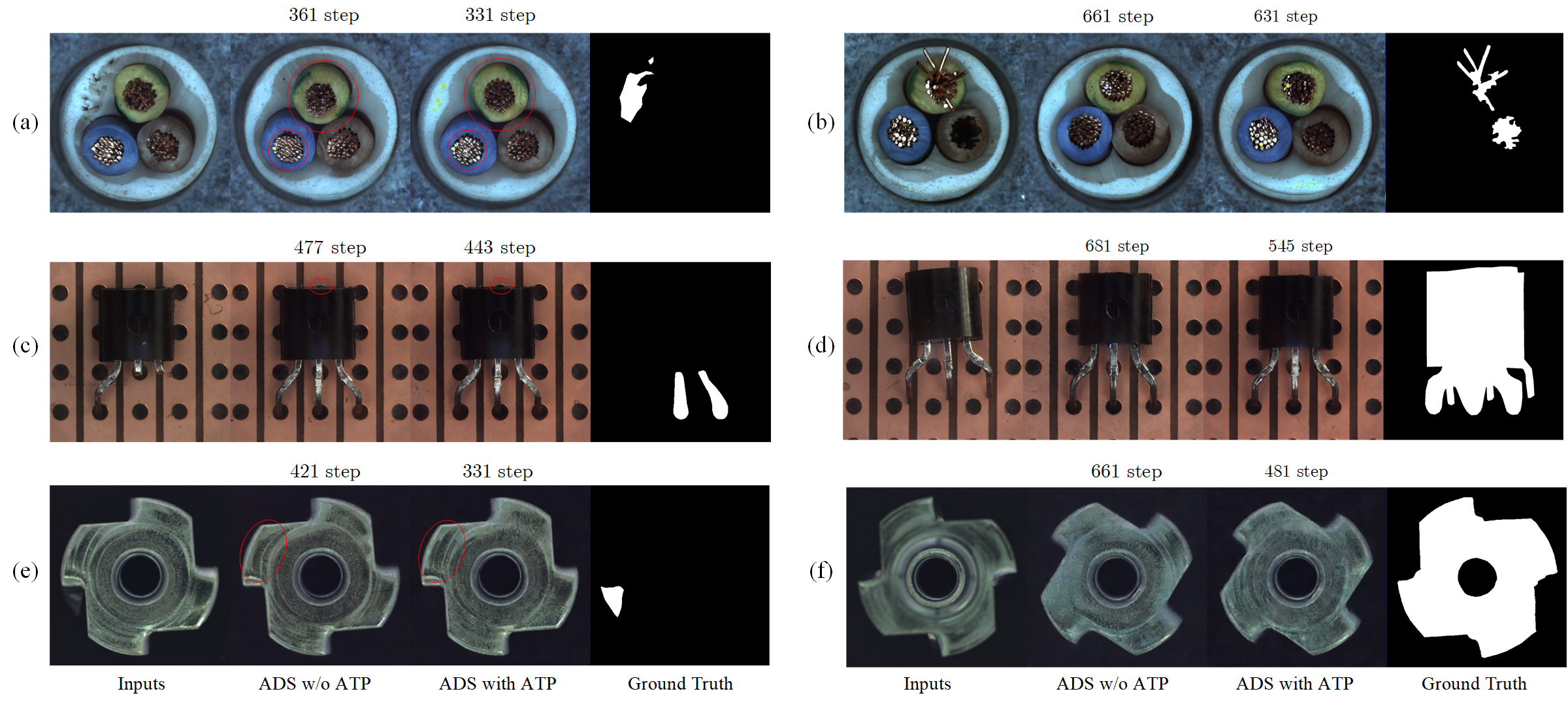}
		\caption{Reconstructions of different types of anomaly and proper steps. Examples (a), (c), and (e) contain small-scale anomalies, and (b), (d), and (f) are large-scale anomalies. The numbers above the reconstructed images represent the proper steps. Differences in details of normal areas are marked in red circles.
		}
		\label{fig:adaptive_step}
	\end{figure}

	We display different types of anomaly and proper steps chosen by ADS in \cref{fig:adaptive_step}. For anomalies that are not obvious, such as examples (a), (c), and (e), ADS tends to select small steps that are enough for reconstruction. However, for large-scale anomalies, such as examples (b), (d), and (f), ADS denoises from samples with larger step noise to ensure the reconstruction of anomaly-free images.
	
	Besides, we observed that ATP can further reduce the denoising steps. \cref{fig:adaptive_step} displays the reconstructions of ADS and the combination of ADS and ATP. Compared with not using ATP, the model can better remove anomalies during the inference process. As a result, a clear image for comparison will contain less anomaly information at each denoising step, and the difference is detected in smaller steps. This allows more normal information to be retained to produce a more accurate reconstruction while ensuring the complete removal of anomalies. Some differences in details are marked with red circles in \cref{fig:adaptive_step}. The reconstruction of ADS with ATP is more similar with inputs at normal areas. Performance can be further improved as shown in \cref{table3}.
	
	\subsubsection{Spatial-Adaptive Feature Fusion}
	\label{sec:ablation_SAFF}
	In Spatial-Adaptive Feature Fusion (SAFF), we fuse the features from the predicted sample and the test sample with a mask $\bm{\mathit{m}}$, which is generated with the anomaly map at the proper denoising step. \cref{table3} shows the effect of SAFF. SAFF can further remove residual anomaly information and improve detection and location performance. 
	
	\subsubsection{Anomaly-oriented Training Paradigm}
\begin{figure}
		\centering
		\includegraphics[width=1.0\linewidth]{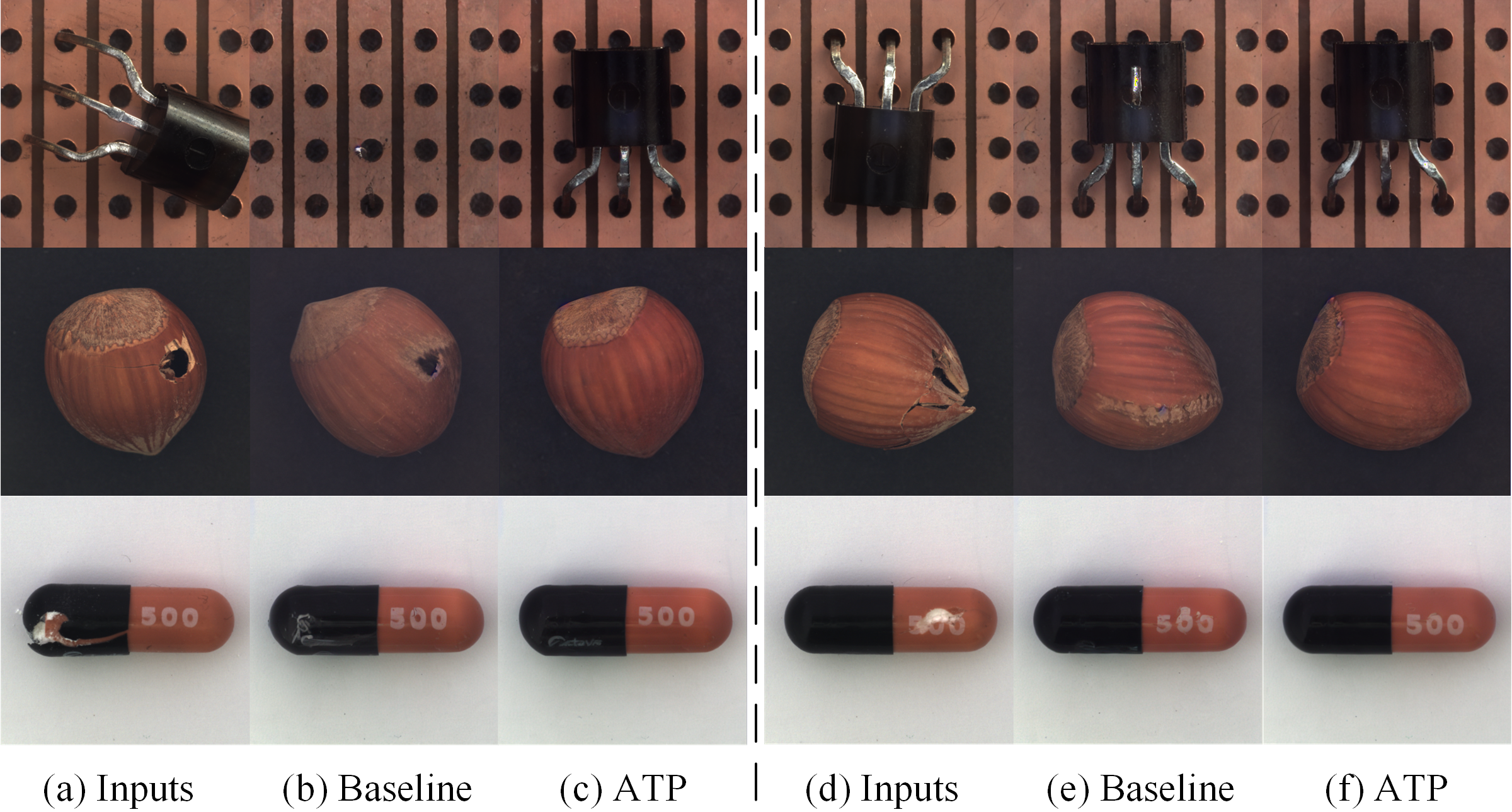}
		\caption{Qualitative comparisons between baseline and proposed ATP on MVTec-AD. The same denoising steps are used for the two methods.
		}
		\label{fig:abonrmal}
	\end{figure}

	Because our training paradigm adds synthesis anomaly samples in training, the data distribution variance between the training data and the test data is narrowed. Thus, DM can map abnormal regions as normal regions as well. Performance comparison of baseline and proposed training paradigm are shown in \cref{table3}, \ie, LDM (baseline) and ATP (proposed anomaly-oriented training paradigm). We also provide visual qualitative comparisons in \cref{fig:abonrmal}. Adaptive denoising steps are not used there. The same steps of noise are added for baseline and ATP. In the (b) column and (e) column, the results of the baseline still contain abnormal areas. This suggests that the reconstruction ability of the baseline is limited. After training with ATP, as shown in the (c) column and (f) column in \cref{fig:abonrmal}, the model can map abnormal regions into normal regions well. 
	
	\section{Social Impact, Limitations, and Future Work}
	This work studies the problem of unsupervised anomaly detection and achieves SOTA performance on four datasets.
	Without the need for real-world abnormal samples, this work has the potential to be efficiently utilized in real-world scenarios.
	Despite the appealing performance, our method introduces evaluation comparison in the inference, which causes extra time costs.
	Besides, to determine the denoising step of our adaptive denoising step strategy, we conduct the comparison step-by-step, and the denoising steps are not actually reduced.
	In future work, we plan to design a lightweight evaluation comparison and try to predict the denoising step with a limited number of denoising steps, which has the potential to improve the efficiency of our method.
	
	\section{Conclusion}
	In this paper, a global and local adaptive diffusion model, GLAD, is presented to improve reconstruction results. As a global configuration, an adaptive denoising step is set for each sample to adapt to different anomalies. Moreover, considering that reconstructing the abnormal regions is different from the normal areas, an adaptive feature fusion scheme is proposed to remove residual anomalies in abnormal regions, and an anomaly-oriented training paradigm is proposed to promote the ability of the diffusion model to reconstruct anomalies into normal regions, which achieve local adaptive reconstructions. We conduct extensive experiments on MVTec-AD, MPDD, VisA, and PCB-Bank datasets. Quantitative and qualitative results evaluate the superiority and effectiveness of our approach.
	
	\section*{Acknowledgement}
	This work was supported in part by the National Key Research and Development Program of China under Grant No. 2023YFA1008500.
	
	%
	%
	
	\bibliographystyle{splncs04}
	\bibliography{main}

	\newpage
	\appendix
	
	\renewcommand{\thefigure}{S\arabic{figure}}
	\renewcommand{\thetable}{S\arabic{table}}
	\renewcommand{\thesection}{S\arabic{section}}
	\renewcommand{\theequation}{S\arabic{equation}}
	
	\setcounter{table}{0}
	\setcounter{figure}{0}
	\setcounter{section}{0}
	\setcounter{equation}{0}
	
	The content of this supplementary material is organized as follows:
	
	\begin{itemize}
		\item More Implementation Details in \cref{sec:more_details}.
		\item Effect of Different Thresholds in \cref{sec:threshold_ads}.
		\item Ablation of Spatial-Adaptive Feature Fusion in \cref{sec:ablation_SAFF_supp}.
		\item Effect of Different Feature Extractors in \cref{sec:feature_extractor}.
		\item Denoising Process with Adaptive Denoising Steps in \cref{sec:adaptive_denoising_step}.
		\item Results at multi-category settings in \cref{sec:multi-category_results}.
		\item Additional Qualitative Results in \cref{sec:additional_qualitative_results}.
	\end{itemize}
	
	\section{More Implementation Details}
	\label{sec:more_details}
	Following DDAD~\cite{DINO39}, We fine-tune the DINO for the VisA and PCB-Bank datasets because of the larger differences between the datasets and the pre-training datasets. Specifically, we optimize DINO by reducing the cosine similarity between the normal multi-layer features $\ii{F_t}$ and the corresponding reconstructed multi-layer feature $\ii{F_r}$. The cosine similarity is calculated as a loss $\ii{L_{FT}}$:
	\begin{equation}
		\begin{split}
			\ii{L_{FT}}& = \ii{L}_{Similarity}(\ii{F_t}, \ii{F_r}) + \lambda \ii{L}_{DL}(\ii{F_t}, \ii{F_r}) \\
			& =         \sum_{l \in J} (1 - \cos(\ii{F^l_t}, \ii{F^l_r})) \\
			& + \lambda (\sum_{l \in J} (1 - \cos(\ii{F^l_t}, \bar{\ii{F^l_t}})) ) \\
			& + \lambda (\sum_{l \in J} (1 - \cos(\ii{F^l_r}, \bar{\ii{F^l_r}})) )
		\end{split}
		\label{eqn:fine-tuning loss}
	\end{equation}
	$J$ is set as $\{3, 6, 9, 12\}$. $\ii{L}_{DL}$ denotes a distillation loss to constrain features from training feature extractor and frozen feature extractor, to mitigate the reduction in generalization of the network. $\bar{\ii{F^l_t}}$ and $\bar{\ii{F^l_r}}$ are layer $l$ features of frozen feature extractor. $\lambda$ is set as 0.01. Learning rate and batch size are $3 \times 10^{-4}$ and 16. Noise of $t_{min}$ step (200 step) is added on training images for reconstruction. We only fine-tuning MLP of layer 3, 6, 9 and 12 of DINO.  
	
	We also test the performance of GLAD at multi-category settings. $T$ is set as 650 for the MVTec-AD dataset, 500 for the MPDD dataset, and 500 for the VisA and PCB-Bank datasets. The minimum step $t_{min}$ is set as 350 for the MVTec-AD dataset and MPDD dataset, and 250 for the VisA and PCB-Bank dataset to avoid missing anomalies. The thresholds are 0.45 for MVTec-AD, 0.35 for MPDD, 0.15 for VisA and 0.2 for PCB-Bank.
	
	\begin{figure}
		\centering
		\includegraphics[width=1.0\linewidth]{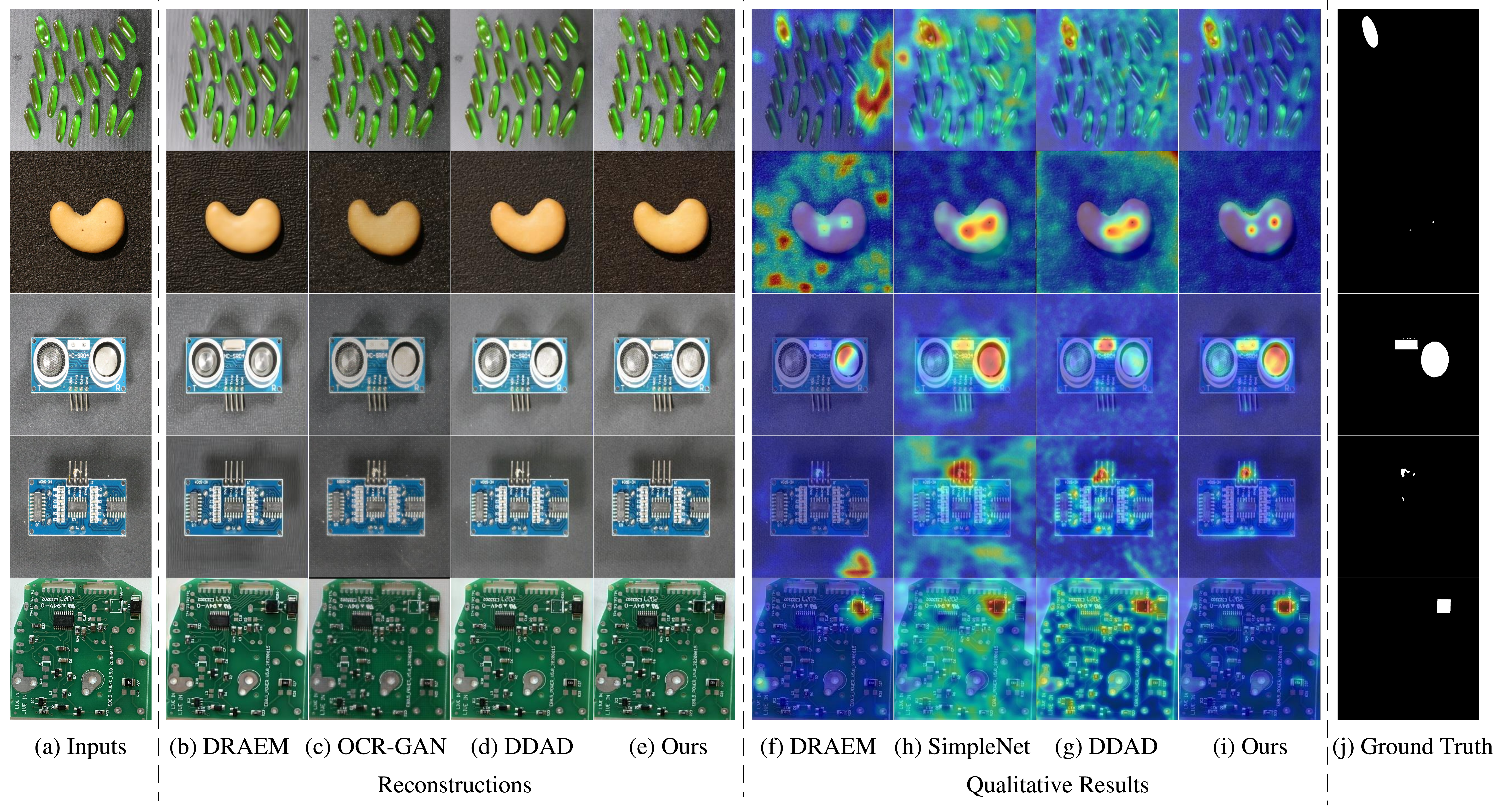}
		\caption{Reconstructions and qualitative results on VisA and PCB-Bank datasets. The first four rows display examples of the VisA dataset, and the last row is for the PCB-Bank dataset.
		}
		\label{fig:heatmap_visa}
	\end{figure}
	
	\section{Effect of Different Thresholds}
	\label{sec:threshold_ads}
	
	\begin{table}
		\caption{Thresholds for different classes on MVTec-AD dataset.}
		\label{table:threshold_mvtec-ad}
		\centering
		\resizebox{1.0\columnwidth}{!}{
			\tabcolsep=0.2cm
			\begin{tabular}{c|cccccccc}
				\toprule
				Class &Carpet &Grid &Leather &Tile &Wood &Bottle &Cable &Capsule \\
				\midrule
				Threshold &0.32 &0.47 &0.35 &0.35 &0.37 &0.32 &0.40 &0.40 \\
				\bottomrule
				Class &Hazelnut &Metal nut &Pill &Screw &Toothbrush &Transistor &Zipper &\\
				\midrule
				Threshold &0.50 &0.40 &0.35 &0.32 &0.50 &0.50 &0.35 &\\
				\bottomrule
			\end{tabular}
		}
	\end{table}
	
	\begin{table}
		\caption{Thresholds for different classes on MPDD dataset.}
		\label{table:threshold_mpdd}
		\centering
		\resizebox{1.0\columnwidth}{!}{
			\tabcolsep=0.2cm
			\begin{tabular}{c|cccccc}
				\toprule
				Class &Bracket Black &Bracket Brown &Bracket White &Connector &Metal Plate &Tubes \\
				\midrule
				Threshold &0.35 &0.35 &0.35 &0.35 &0.25 &0.10 \\
				\bottomrule
			\end{tabular}
		}
	\end{table}
	
	\begin{table}
		\caption{Thresholds for different classes on VisA dataset.}
		\label{table:threshold_visa}
		\centering
		\resizebox{1.0\columnwidth}{!}{
			\tabcolsep=0.3cm
			\begin{tabular}{c|cccccc}
				\toprule
				Class &Candle &Capsules &Cashew &Chewinggum &Fryum  &Macaroni1 \\
				\midrule
				Threshold &0.45 &0.40 &0.40 &0.45 &0.35 &0.45 \\
				\bottomrule
				Class &Macaroni2 &Pcb1 &Pcb2 &Pcb3 &Pcb4 &Pipe fryum\\
				\midrule
				Threshold &0.45 &0.30 &0.30 &0.30 &0.30 &0.45 \\
				\bottomrule
			\end{tabular}
		}
	\end{table}
	
	\begin{table}	
		\caption{Thresholds for different classes on PCB-Bank dataset.}
		\label{table:threshold_pcbbank}
		\centering
		\resizebox{1.0\columnwidth}{!}{
			\tabcolsep=0.5cm
			\begin{tabular}{c|ccccccc}
				\toprule
				Class &Pcb1 &Pcb2 &Pcb3 &Pcb4 &Pcb5 &Pcb6 &Pcb7 \\
				\midrule
				Threshold &0.30 &0.30 &0.30 &0.30 &0.40 &0.45 &0.30 \\
				\bottomrule
			\end{tabular}
		}
	\end{table}
	
	\begin{figure}
		\centering
		\includegraphics[width=1.0\linewidth]{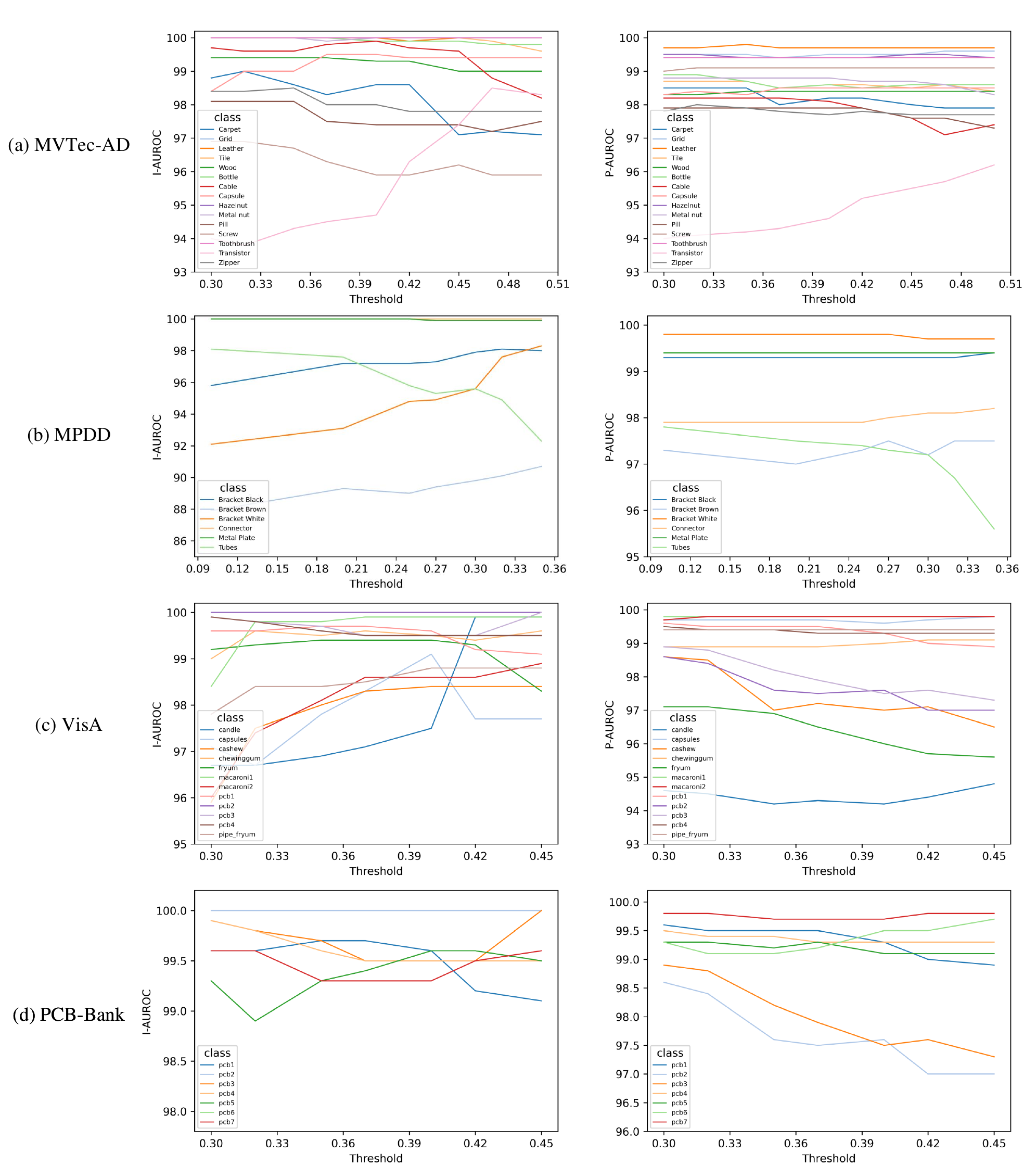}
		\caption{Performance with different thresholds $\delta$. (a), (b), (c), (d) denote the results for MVTec-AD, MPDD, VisA and PCB-Bank datasets, respectively.
		}
		\label{fig:threshold}
	\end{figure}
	
	In our method, threshold is used to measure whether the different between reconstructed sample and noisy input is enough large to determine proper steps. A large threshold may result in that model denoises from a small step and anomaly may be retained in reconstructed image. On the contrary, a small threshold makes it difficult for the model to retain normal information. We display the results of different thresholds in \cref{fig:threshold}. The thresholds are experimentally selected as \cref{table:threshold_mvtec-ad} for MVTec-AD and \cref{table:threshold_mpdd} for MPDD, \cref{table:threshold_visa} for VisA and \cref{table:threshold_pcbbank} for PCB-Bank datasets. In \cref{fig:threshold} (a), as the threshold increases, the performance of transistor changes rapidly. Different from other classes, there are cluttered backgrounds in images of transistor. A larger threshold, which means smaller denoising steps, leads to less background difference and better performance. Thus, the larger thresholds significantly improve performance. For class tubes in MPDD dataset, the background is relatively simple, but anomaly is easy to ignored with a large threshold. And anomalies may be preserved in reconstructed images. Thus, the increased threshold results in a sharp drop in performance.

	\section{Ablation of Spatial-Adaptive Feature Fusion}
	\label{sec:ablation_SAFF_supp}
	In Spatial-Adaptive Feature Fusion (SAFF), we fuse the features from the predicted sample and the test sample with a mask $\bm{\mathit{m}}$, which is generated with the anomaly map at proper denoising step. We test different ways to generate mask in SAFF, including SAFF-norm (anomaly map subtracts its minimum value, then divides it by the difference between the maximum value and the minimum value), and SAFF-sigmod (using sigmod). \cref{table:SAFF} shows the effect of SAFF and different generation ways of the mask. In experiments, SAFF with sigmod achieves best results and is default settings in our method.
	\begin{table}[h]
		\caption{Results of different SAFF.}
		\label{table:SAFF}
		\centering
		\begin{tabular}{c|cccccc}
			\toprule
			Methods &I-AUROC &P-AUROC \\
			\midrule
			W/o SAFF           &99.2 &98.3 \\
			SAFF-norm &99.2 &\textbf{98.6}\\
			SAFF-sigmod(Ours) &\textbf{99.3} &\textbf{98.6}\\
			\bottomrule
		\end{tabular}
	\end{table}
	
	\section{Effect of Different Feature Extractors}
	\label{sec:feature_extractor}
	\cref{table:feature_extractors} shows the results of different pre-trained feature extractors. Following AprilGAN\cite{APRILGAN18}, which uses CLIP\cite{clip1}, we first test the CLIP. In addition, we also tested the DINO\cite{DINO39}, which is a self-supervised vision transformers. Compared with CLIP, DINO achieves better results because of more excellent visual feature extraction capability. Besides, a phenomenon can be observed that smaller patch size, which means larger feature maps and carries more information, obtains better results.
	\begin{table}[h]
		\caption{Results of different pre-trained feature extractors.}
		\label{table:feature_extractors}
		\centering
		\tabcolsep=0.2cm
		\begin{tabular}{c|cccccc}
			\toprule
			Feature Extractors &Architectures &I-AUROC &P-AUROC \\
			\midrule
			\multirow{2}{*}{CLIP}    &ViT-B/16 &95.1 &95.3 \\
			&ViT-L/14 &97.8 &96.3 \\
			\midrule
			\multirow{2}{*}{DINO}   
			&ViT-B/16 &97.7 &98.4\\
			&ViT-B/8  &\textbf{99.3} &\textbf{98.6}\\
			\bottomrule
		\end{tabular}
	\end{table}
	
	\section{Denoising Process with Adaptive Denoising Steps}
	\label{sec:adaptive_denoising_step}
	
	\begin{figure}
		\centering
		\includegraphics[width=1.0\linewidth]{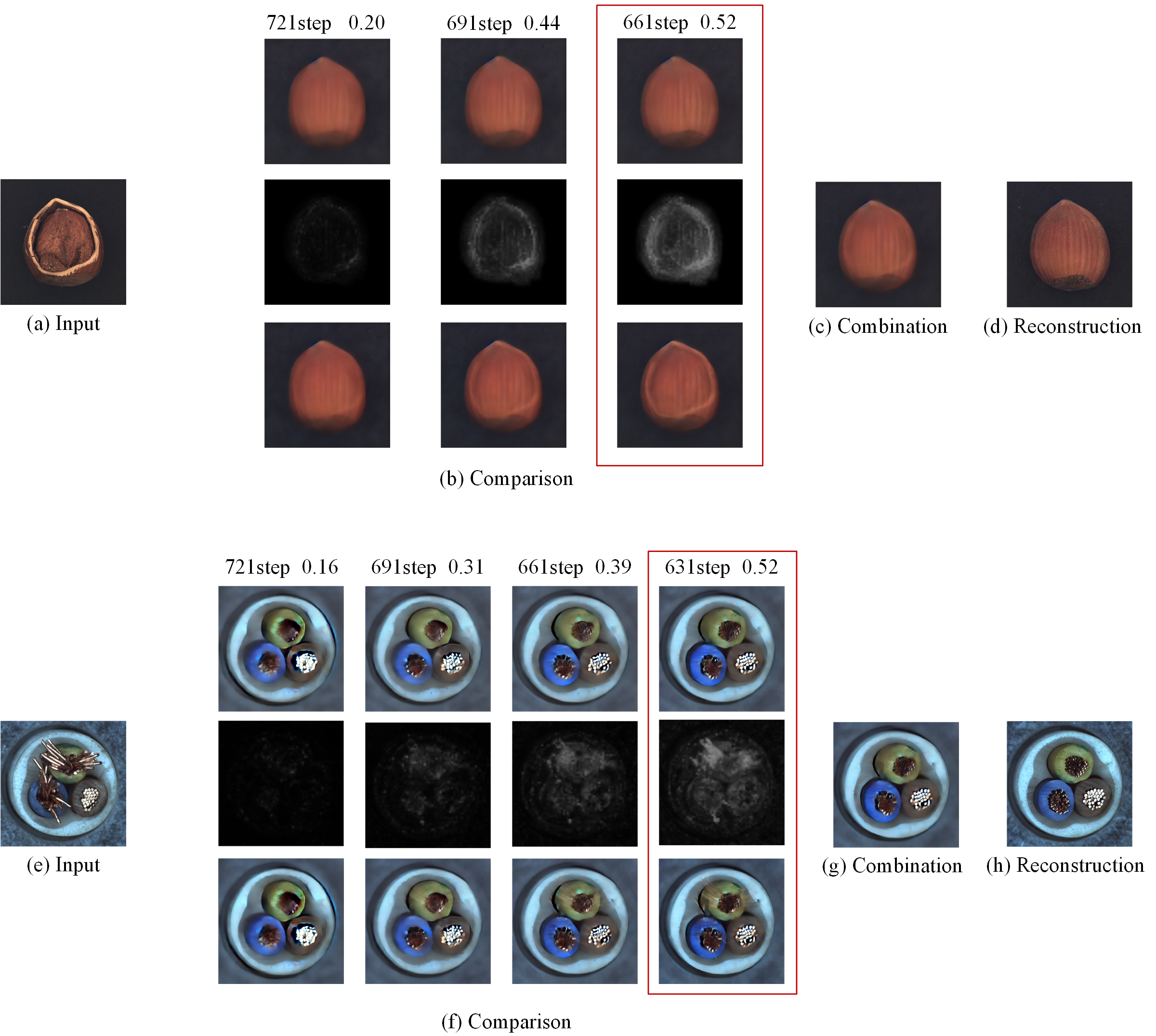}
		\caption{Examples of adaptive step process. Abnormal images are input as (a) and (c). In comparison (b) and (f), the numbers at the top of the images represent the current denoising steps and anomaly scores, and the images from first to third rows are clear images from predicted samples, anomaly maps and clear images from test samples.The steps masked by red boxes denote proper steps, at which, (c) and (g) are composed for next denoising steps. Finally, (d) and (h) are reconstructed anomaly-free images.
		}
		\label{fig:adptive_step_exp}
	\end{figure}
	
	In this section, we demonstrate two examples of denoising process with adaptive steps in \cref{fig:adptive_step_exp}. Test images are inputted and comparison is conducted in each step to search the proper steps. For example (a) of hazelnut, the different is larger than threshold (0.52 > 0.5) at 661 step, which is consider as the proper step. Then the feature are combined as (c) according to the anomaly map of 661 step. Finally, the anomaly-free counterpart (d) is reconstructed. The example (e) of cable goes through the same process, except that the proper step is 631.
	
	\section{Results at Multi-category Settings}
	\label{sec:multi-category_results}
	\cref{table:more_exp_multi-class} shows the quantitative results of multi-category setting. Our proposed GLAD outperforms the latest SOTA on all metrics, and exceeds DiAD on average results of 4 datasets by 6.3↑/5.6↑/4.9↑/1.9↑/14.3↑/10.4↑/11.1↑ on I-AUROC/I-AP/I-F1-max/P-AUROC/P-AP/P-F1-max/PRO.

	\begin{table}
		\caption{Quantitative results of multi-category setting on MVTec-AD, MPDD, VisA and PCB-Bank datasets. Metrics are I-AUROC/I-AP/I-F1-max at first raw (for detection) and P-AUROC/P-AP/P-F1-max/PRO at second raw (for localization).}
		\centering
		\label{table:more_exp_multi-class}
		\resizebox{1.0\columnwidth}{!}{
			\begin{tabular}{ccccc|c}
				\hline
				\makebox[0.1\textwidth][c]{Dataset}   &\makebox[0.25\textwidth][c]{MVTec-AD}  & \makebox[0.25\textwidth][c]{MPDD}               &\makebox[0.25\textwidth][c]{VisA}           & \makebox[0.25\textwidth][c]{PCB-Bank}  &\makebox[0.25\textwidth][c]{Avg}     \\      
				\hline
				
				\multirow{2}{*}{UniAD~\cite{UniAD6}}      &96.5/98.8/96.2      &82.5/85.9/85.9       &85.5/85.5/84.4          & 86.1/85.2/84.1      &87.7/88.9/87.7 \\
				&96.8/43.4/49.5/90.4 &95.1/18.5/25.0/81.7 &95.9/21.0/27.0/75.6      &97.2/19.4/26.4/78.8          &96.3/25.6/32.0/81.6\\
				
				\multirow{2}{*}{DiAD~\cite{DiAD37}}        &97.2/99.0/96.5       &82.8/84.9/85.8       &86.8/88.3/85.1          &89.0/88.8/85.2      &89.0/90.3/88.2\\
				&96.8/52.6/55.5/90.7  &94.8/19.4/25.4/82.7  &96.0/26.1/33.0/75.2     &96.9/24.5/33.6/79.4  &96.1/30.7/36.9/82.0\\  
				\hline 
				
				\multirow{2}{*}{Ours-256}    &\textbf{97.5/99.1/96.6}       &\textbf{97.5/97.1/96.8}      &\textbf{91.8/92.9/88.6}           &\textbf{94.2/94.3/90.2}        &\textbf{95.3/95.9/93.1}\\
				&\textbf{97.4/60.8/60.7/93.0}  &\textbf{98.0/40.9/41.5/93.0}  &\textbf{97.8/35.6/40.9/92.1}       &\textbf{98.8/42.6/46.1/94.1} &\textbf{98.0/45.0/47.3/93.1} \\ 
				
				\bottomrule   
			\end{tabular}
		}
	\end{table}
	
	\section{Additional Qualitative Results}
	\label{sec:additional_qualitative_results}
	
\begin{figure}
		\centering
		\includegraphics[width=1.0\linewidth]{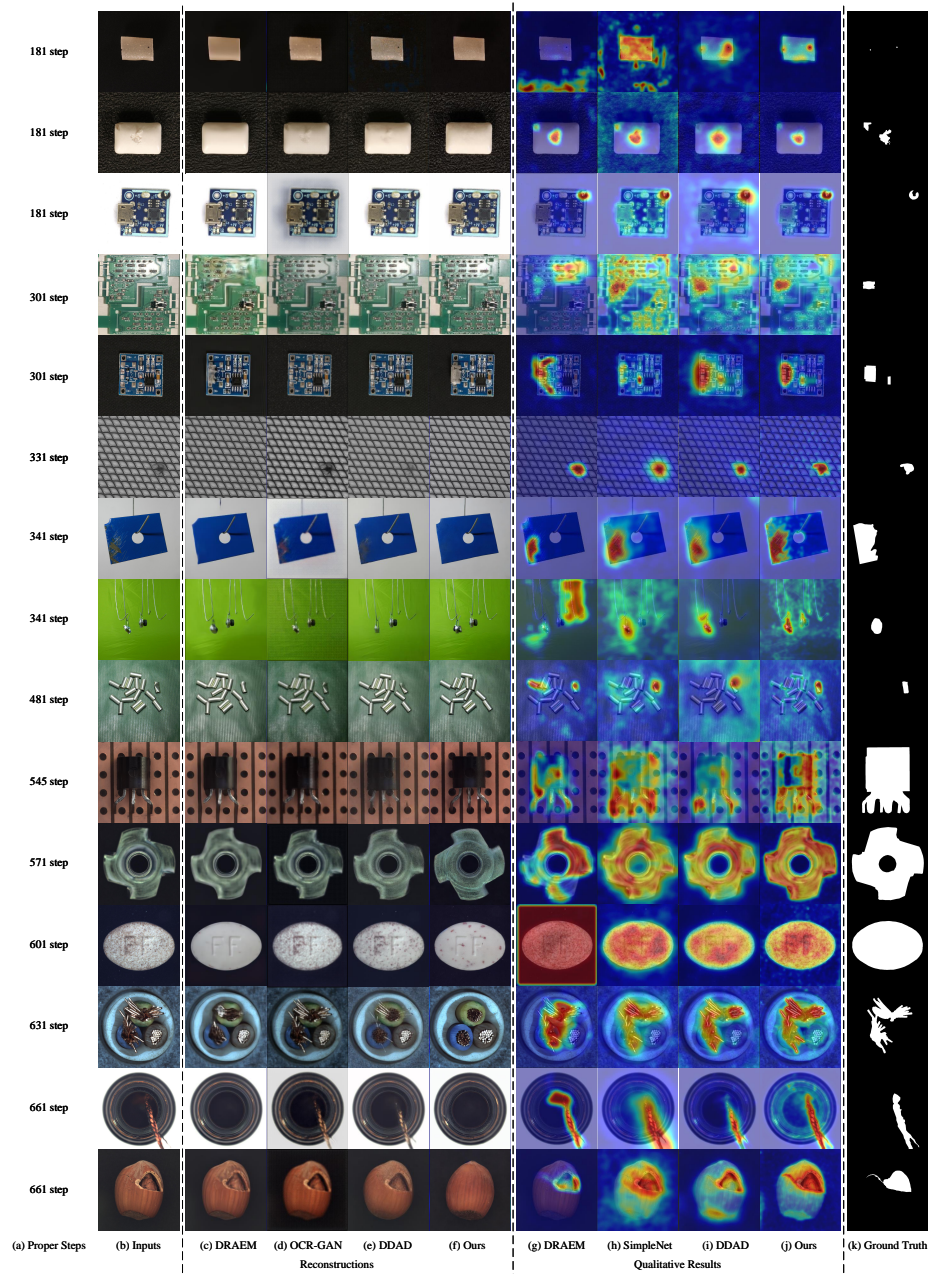}
		\caption{Additional reconstructions and qualitative results on MVTec-AD, MPDD, VisA and PCB-Bank datasets. The difficulty of sample reconstruction from top to bottom gradually increases, and the corresponding denoising steps are shown in column (a).
		}
		\label{fig:heatmap_supp}
	\end{figure}

	Some additional reconstructions and qualitative results are displayed in \cref{fig:heatmap_supp}. First for reconstruction, our method can achieve satisfactory reconstruction results. Even for some large-scale anomalies, our method can still generate anomaly-free images, such as transistor, metal nut, cable and hazelnut. In terms of qualitative results, compared with other methods, our method enables more accurate and refined localization. All results show the superiority of our method.
	
\end{document}